\documentclass{article}

\usepackage{microtype}
\usepackage{graphicx}
\usepackage{subfigure}
\usepackage{booktabs} % for professional tables

\usepackage{hyperref}

\usepackage[accepted]{icml2024}

\usepackage{amsmath}
\usepackage{amssymb}
\usepackage{mathtools}
\usepackage{amsthm}
\usepackage{listings}
\usepackage{fancyvrb}

\usepackage[inline]{enumitem}

\usepackage[capitalize,noabbrev]{cleveref}

\theoremstyle{plain}

\theoremstyle{definition}

\theoremstyle{remark}

\usepackage[textsize=tiny]{todonotes}

\icmltitlerunning{Adversarial Robustness Limits}

\usepackage{multirow}
\usepackage{xcolor}
\usepackage{tcolorbox}
\tcbuselibrary{skins}

\DeclareMathOperator*{\maximize}{max}
\DeclareMathOperator*{\minimize}{min}
\DeclareMathOperator{\proj}{proj}
\DeclareMathOperator{\sign}{sign}
\usepackage{widetable}
\usepackage[group-separator={,}]{siunitx}

\begin{document}

\twocolumn[
%\icmltitle{Adversarial Alignment Precedes Adversarial Robustness and Coincides with  Robustness Scaling Law Asymptote}
%\icmltitle{The Adversarial Misalignment Limit Makes Adversarial Robustness Unsolvable and Is Consistent with Scaling Law Asymptote}
%\icmltitle{Why is adversarial robustness unsolved? Scaling laws \& an alignment hypothesis}
%\icmltitle{The Adversarial Misalignment Limit Implies Scale Cannot Solve the Adversarial Robustness Problem and Is Consistent with Scaling Law Asymptotes}
%\icmltitle{The Adversarial Misalignment Limit Explains Scaling's Inability to Solve Adversarial Robustness}
%\icmltitle{Why is CIFAR10 adversarial robustness unsolved? Scaling laws vs. alignment}
%\icmltitle{Understanding Adversarial Robustness Requires Rethinking Adversarial Data}
%\icmltitle{Attainable Goals for Adversarial Robustness and Progress towards Them via Scaling-Law and Human-Alignment Studies}
%\icmltitle{Limits to Adversarial Robustness and Progress Towards Them via Scaling-Law and Human-Alignment Studies}
\icmltitle{Adversarial Robustness Limits via Scaling-Law and Human-Alignment Studies}
%\icmltitle{A Human Aligned Roofline Scaling Law: FLOPS versus Adversarial Robustness Accuracy}



\icmlsetsymbol{equal}{*}

\begin{icmlauthorlist}
\icmlauthor{Brian R. Bartoldson}{lab}
\icmlauthor{James Diffenderfer}{lab}
\icmlauthor{Konstantinos Parasyris}{lab}
\icmlauthor{Bhavya Kailkhura}{lab}
\end{icmlauthorlist}

\icmlaffiliation{lab}{Lawrence Livermore National Laboratory}
%\icmlaffiliation{comp}{Company Name, Location, Country}
%\icmlaffiliation{sch}{School of ZZZ, Institute of WWW, Location, Country}

\icmlcorrespondingauthor{Brian and Bhavya}{bartoldson@llnl.gov, kailkhura1@llnl.gov}
%\icmlcorrespondingauthor{Firstname2 Lastname2}{first2.last2@www.uk}

\icmlkeywords{Machine Learning, ICML}

\vskip 0.3in
]



\printAffiliationsAndNotice{}  % leave blank if no need to mention equal contribution
%\printAffiliationsAndNotice{} % otherwise use the standard text.
%\printAffiliationsOnly{} % BRIAN ADDED THIS, DO NOT USE FOR FINAL VERSION, JUST FOR PREPRINT

\begin{abstract}
    
%Subtle changes to neural net input can cause undesirable behaviors ranging from image misclassification to generative modeling guardrail-failure. Robustness to these failures is improved by training on adversarially perturbed data, but the limits of robustness are not well understood. 
%On CIFAR10 for instance, SOTA accuracy is about $100$\%, but SOTA robustness barely exceeds $70$\% despite use of hundreds of millions of synthetic training samples and over $10^{21}$ FLOPs, almost a tenth of the Llama 7B pretraining cost. 
%Specifically, we study how CIFAR10 robustness responds to changes in three key factors---model size, dataset size, and synthetic data quality---by fitting the first scaling laws that model their effects to hundreds of training runs. 

%Our scaling laws accurately predict robustness, allowing us to match the prior SOTA with $20$\% ($55$\%) fewer training (inference) FLOPs and to establish a new SOTA ($74$\%, $+3$\% AutoAttack accuracy) with more FLOPs.
%However, our scaling laws also predict robustness slowly climbs and plateaus at 90\%: dwarfing our SOTA by scaling is impractical, and perfect robustness is impossible. 
This paper revisits the simple, long-studied, yet still unsolved problem of making image classifiers robust to imperceptible perturbations. Taking CIFAR10 as an example, SOTA clean accuracy is about $100$\%, but SOTA robustness to $\ell_{\infty}$-norm bounded perturbations barely exceeds $70$\%. To understand this gap, we analyze how model size, dataset size, and synthetic data quality affect robustness by developing the first scaling laws for adversarial training. Our scaling laws reveal inefficiencies in prior art and provide actionable feedback to advance the field. For instance, we discovered that SOTA methods diverge notably from compute-optimal setups, using excess compute for their level of robustness. Leveraging a compute-efficient setup, we surpass the prior SOTA with $20$\% ($70$\%) fewer training (inference) FLOPs. We trained various compute-efficient models, with our best achieving $74$\% AutoAttack accuracy ($+3$\% gain). However, our scaling laws also predict robustness slowly grows then plateaus at $90$\%: dwarfing our new SOTA by scaling is impractical, and perfect robustness is impossible. To better understand this predicted limit, we carry out a small-scale human evaluation on the AutoAttack data that fools our top-performing model. Concerningly, we estimate that human performance also plateaus near $90$\%, which we show to be attributable to $\ell_{\infty}$-constrained attacks' generation of invalid images not consistent with their original labels. Having characterized limiting roadblocks, we outline promising paths for future research.

\end{abstract}

\section{Introduction}
\label{intro}

\begin{figure*}[t]
    \centering
    \includegraphics[width=1\linewidth]{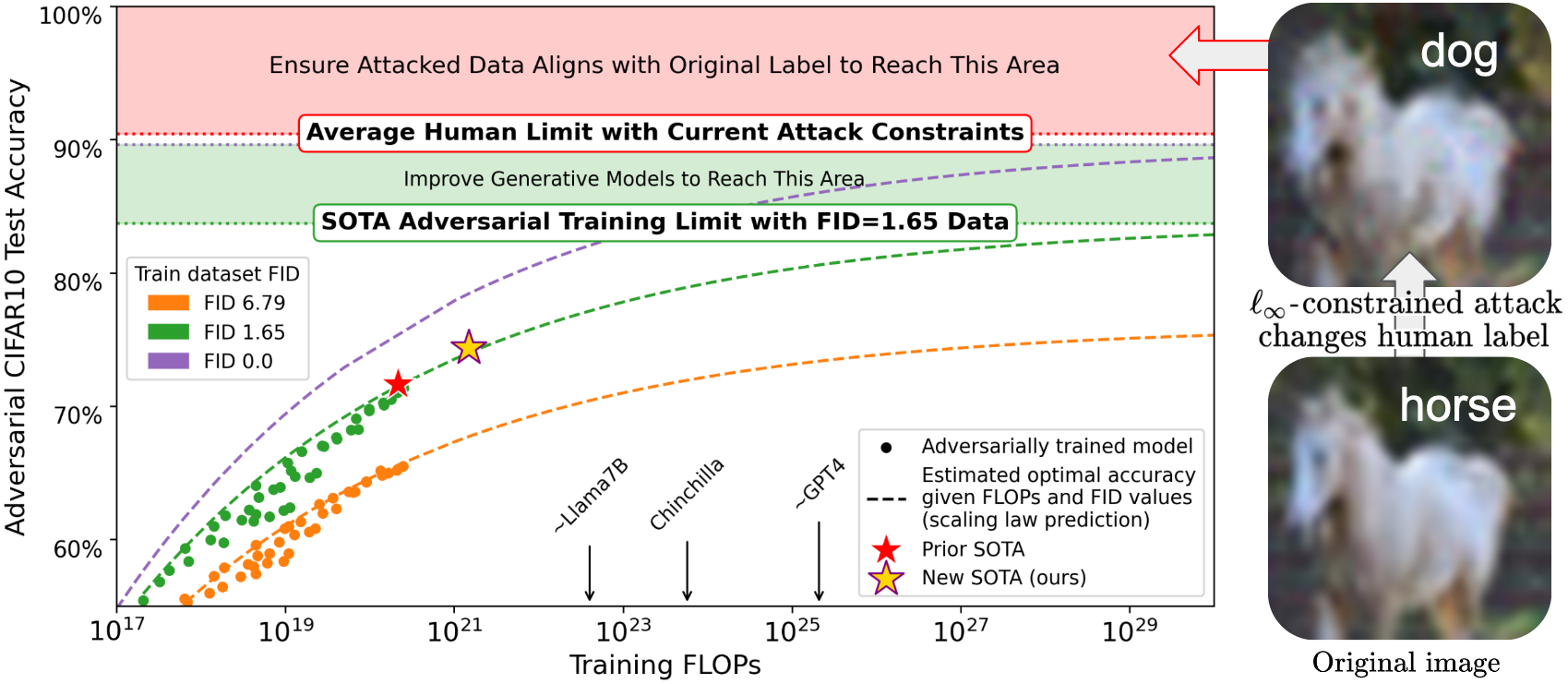}
    %\vspace*{-7mm}
    \caption{\textbf{SOTA techniques cannot solve the CIFAR10 adversarial robustness problem, even with infinite compute.} (\textbf{left}) Scaling laws predict that performance asymptotes near 90\%. We produce this estimate using Approach 2 (Section \ref{sec:approach_2}), a scaling law approach we designed that can model the effect of FID, allowing approximation of performance when training on a hypothetical FID=0.0 synthetic dataset of arbitrarily large size. Note that lowering FID raises dataset quality. We learned the parameters of Approach 2's parametric model of testing performance by fitting to the performances that resulted from training various NNs on various datasets; the depicted dots show a subset of these performances. (\textbf{right}) A CIFAR10 test image is adversarially perturbed by AutoAttack, causing humans to disagree with the ground-truth label ``horse''. Consistent with scaling law asymptotes, our small-scale human study (Section \ref{sec:human}) estimates that the original label is no longer an appropriate label for roughly $10$\% of adversarial data. Benchmarking that uses the original label after it ceases to be a good fit for the image will suggest that NN robustness is further from being ``solved'' (human-level) than it truly is.}
    \label{fig:teaser}
\end{figure*}

Neural networks match human performance on several tasks requiring processing of vision and text data \citep{achiam2023gpt,team2023gemini}. However, model performances typically falter when data is adversarially perturbed by attacks humans are robust to \citep{szegedy2013intriguing}, calling into question model trustworthiness. %, safety, and deployability. 
Progress on this problem has been made through adversarial training \citep{goodfellow2014explaining, madry2017towards}, but even on the ``toy'' dataset CIFAR10, the best robustness to $\ell_{\infty}$ attacks is just 71\% according to RobustBench \citep{croce2020robustbench}. 

The absence of a strong adversarial defense for a simple dataset like CIFAR10 is concerning because it suggests that robustifying foundation models may be impractical or even impossible \citep{zou2023universal,bailey2023image,jain2023baseline}. Therefore, we aim to understand, \textit{``Why is the CIFAR10 adversarial robustness problem unsolved?''} 

Our experiments consider the possibility that the answer is a lack of scale used by state-of-the-art methods \citep{wang2023better}. Yet, we find that scale is not enough to achieve robustness: even with unlimited CIFAR10 data, we estimate that reaching human performance would require roughly $10^{30}$ FLOPs (see Figure \ref{fig:teaser}), about 3,000 years of TF32 matrix math on 25,000 MI300 or H100 GPUs. 
Therefore, meaningful progress on this problem requires the design of more efficient training algorithms and improved architectures, rather than simply scaling.

Surprisingly, even if human performance could be reached, our analysis suggests that this problem will still be ``unsolved'': We find that human performance is around 90\% on AutoAttack data (see Figure \ref{fig:teaser}), with the $\sim$$10$\% error induced by \textit{invalid adversarial data} that no longer fits its label. Thus, attack formulations must be rethought to make this problem solvable; e.g., attacks should only produce valid images that abide by the original label.

Our contributions (see Appendix \ref{sec:contributions}) are broadly as follows:\footnotetext[0\def\thefootnote{}]{See code or take online quiz via \url{https://github.com/bbartoldson/Adversarial-Robustness-Limits}.}

%Broadly, our major contributions are as follows\footnote{Appendix~\ref{sec:contributions} provides a detailed list of our contributions and their implications, with links to relevant sections to aid navigation.}:

%\footnote{A more detailed list of contributions and their implications are provided in Appendix~\ref{sec:contributions} to assist readers in navigating the paper and understanding its key points more efficiently.}:
\iffalse
\begin{enumerate}[labelindent=0pt, labelwidth=!, wide]
    \item We derive robustness scaling laws that model synthetic training dataset quality and find model and dataset sizes should be scaled at more equal rates as data quality rises;
    \item Our scaling laws accurately model robustness and suggest efficient training setups that we use to surpass the prior SOTA's efficiency and robustness ($+3\%$ AutoAttack);
    \item We find human and scaling law robustnesses each asymptote well before $100$\% due to flaws in the $\ell_{\infty}$ attack formulation that allow \textit{invalid adversarial data}, which humans misclassify. We study this data and its effect on benchmarking.
\end{enumerate}
\fi

\begin{enumerate}[labelindent=0pt, labelwidth=!, wide]
    \item We develop scaling laws for adversarial robustness, integrating the data quality of generative models. Our scaling laws can accurately predict robustness of unseen configurations, recommend optimal resource allocations, and identify opportunities to reduce model size (raise inference speed).
    \item By applying actionable guidelines from our scaling laws, compared to the prior SOTA \citep{wang2023better}, we achieve the same robustness level with $20$\% fewer FLOPs, enable a $+1$\% AutoAttack gain with a $3\times$ smaller model trained with the same compute budget, and set a new SOTA ($+3$\% AutoAttack accuracy) using a larger compute budget.   
    \item Our analysis suggests that advancements in generative models, efficient algorithms, and model architectures will help address this problem to some extent but will not solve it completely. We find that the $\ell_{\infty}$ attack formulation is flawed and creates \textit{invalid adversarial data} that humans also misclassify. Therefore, solving this problem requires fixing the attack formulation to account for image validity.
\end{enumerate}

\begin{figure*}[t]
\centering
\includegraphics[width=0.9\textwidth]{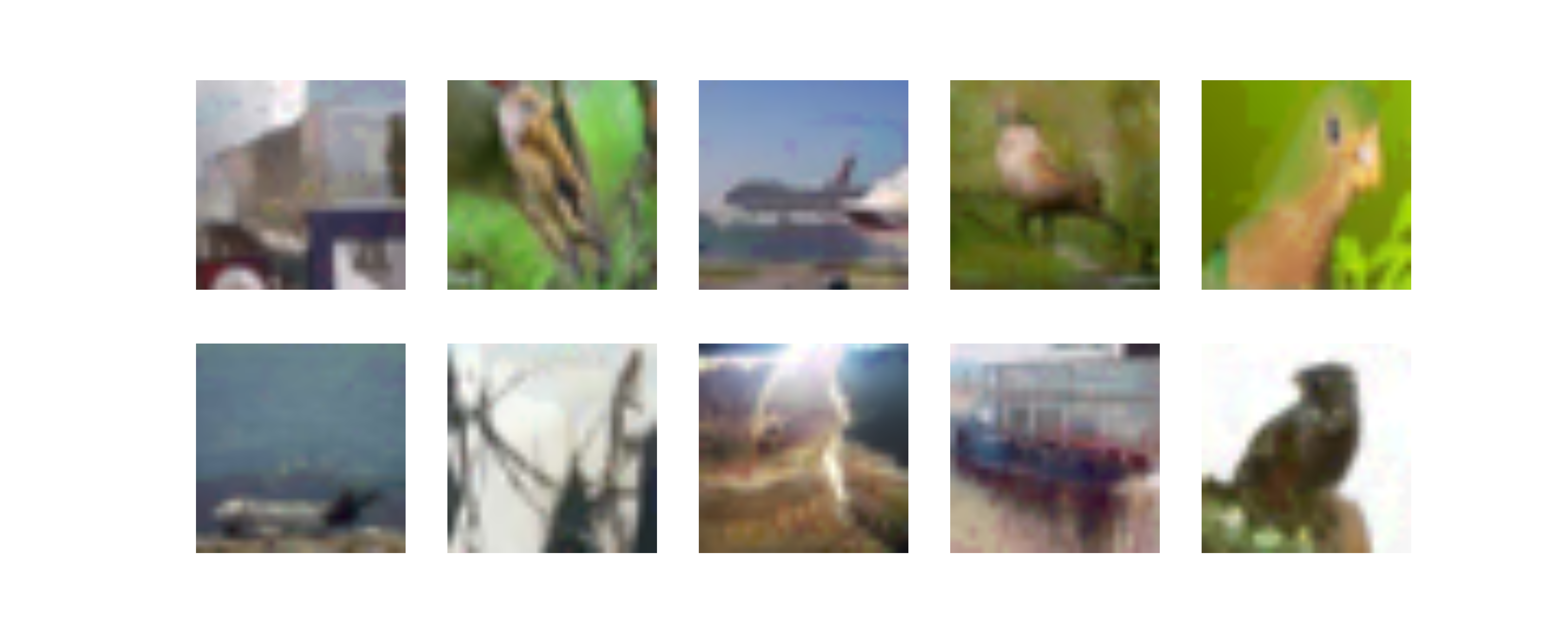}
\vspace{-0.25in}
\caption{\textbf{Subtle attacks transfer to humans, showing brittleness in human visual perception}.
%\caption{\textbf{Exploring the Brittleness of Human Visual Perception}.
%Try to classify these adversarial images, designed to fool our SOTA neural net by exploiting subtle changes, into CIFAR10 classes (airplane, automobile, bird, cat, deer, dog, frog, horse, ship, and truck). Next, see Figure \ref{fig:fool-humans} for the clean counterparts of these images. Were you also fooled?
Try to classify these adversarial images that fool our SOTA neural net as CIFAR10 classes (airplane, automobile, bird, cat, deer, dog, frog, horse, ship, and truck). Next, see Figure \ref{fig:fool-humans} for the clean counterparts of these images to see if you were also fooled.
Section \ref{sec:human} further discusses such ``invalid'' data.
}
\label{fig:adversarial}
\end{figure*}

\section{Related Work}

\paragraph{Adversarial Training Trends}

It is common practice to adversarially train on CIFAR10 data by modifying training data via a 10-step attack \citep{rebuffi2021data,sehwag2021robust,wang2023better,peng2023robust}, making an adversarial training iteration $9\times$ as expensive as a regular one (see Appendix \ref{sec:FLOPs}). Further adding to the costs of adversarial training, recent work find that larger models \citep{singh2023revisiting,huang2023revisiting} and larger datasets improve adversarial robustness. Indeed, synthetic CIFAR10 datasets with as many as 50M samples have been used to boost adversarial training's effectiveness \citep{rebuffi2021data,gowal2021improving,sehwag2021robust,wang2023better}. Our SOTA model trains on 300M unique synthetic CIFAR10 data points, requiring over $10^{21}$ training FLOPs (Figure \ref{fig:teaser}).

Aware of the cost needed just to make CIFAR10 models moderately robust, recent research on safeguarding LLMs and vision language models (VLMs) against attacks discusses the potential difficulty or even futility of robustifying foundation models \citep{zou2023universal,bailey2023image,jain2023baseline, sun2024trustllm}. Before incorporating any adversarial training or robustness measures, the training cost of such models is already high---Figure \ref{fig:teaser} shows the FLOPs costs of training: GPT4 \citep{achiam2023gpt}, a closed model with a rough cost estimate \citep{epoch2023aitrends}; Llama7b \citep{touvron2023llama}, the cost of which we estimate by applying the $\mathrm{FLOPs}=6ND$ heuristic of \citet{kaplan2020scaling} to 1T tokens and 6.7B parameters; and Chinchilla \citep{hoffmann2022training}. Notably, such costs have been tempered by use of compute-efficient training settings derived from scaling laws fit to LLM performances \citep{kaplan2020scaling}. We consider the possibility that similar scaling laws could help address the inefficiency of adversarial training. 

%brian
Given the robustness benefits of adversarial training with large synthetic datasets \citep{rebuffi2021data,gowal2021improving} and with improved synthetic data quality as measured by FID \citep{wang2023better}, our scaling law approach generalizes prior scaling laws by modeling the effect of synthetic data quality. As a result, our scaling law predictions change as training dataset FID changes (see Figure \ref{fig:teaser}), which is not the case in smaller concurrent studies of the scalability of adversarial training \citep{debenedetti2023scaling}. Notably, prior work incorporated data quality into language model scaling laws by decreasing the contribution of duplicative training data \citep{muennighoff2023scaling}, whereas we do not train on duplicate training data (except where noted) and model quality using FID. When data quality is sufficiently high, we find that efficient adversarial training scales the model size and dataset size at roughly the same rate, consistent with Chinchilla's scaling laws \citep{hoffmann2022training}.

\paragraph{Adversarial Examples That Fool Humans} 

In Figures \ref{fig:teaser} and \ref{fig:adversarial}, we show that running AutoAttack \citep{croce2020reliable} on our SOTA NN produces adversarial data
%---with a maximum pixel perturbation $\ell_{\infty}$ norm of $8/255$---
that tricks both humans and NNs.
%into predicting that the image contains a dog. 
Prior work shows humans are also fooled by adversarial data with higher resolutions and different perturbation-norm constraints \citep{elsayed2018adversarial,guo2022adversarially,veerabadran2023subtle,gaziv2023robustified}. 
For instance, small-$\ell_{2}$ norm  adversarial perturbations, which are often discussed as mostly unnoticeable by humans, can fool humans on high-resolution ImageNet data \citep{gaziv2023robustified}. 
Importantly, \citet{gaziv2023robustified} find that humans are more easily tricked by adversarial data if the network that's attacked to generate the data is first robustified.

Models not trained to be robust can also be used to generate adversarial data that fool humans, though humans are only fooled by such images when given limited time to classify them \citep{elsayed2018adversarial}.  In \citet{gaziv2023robustified} and in this work, adversarial data is generated using robust networks, and humans may not deduce the correct category even with extra time. The ability to trick humans using images that trick robust neural networks may not be surprising given the overlap in frequencies relied upon by each system to make classifications \citep{subramanian2023spatial},
\footnote{Humans use a subset of NN frequencies (see Appendix \ref{sec:tricking_humans}).} 
%\footnote{Humans actually use a subset of the frequencies used by neural networks. See Appendix \ref{sec:tricking_humans} for more detailed discussion.} 
the potential for a shared mechanism for robustness in humans and machines \citep{harrington2021finding}, and the susceptibility of biological, primate neurons to adversarial attacks \citep{guo2022adversarially}. However, we identify that such human susceptibility causes benchmarking of robustified networks \citep{croce2020robustbench} to produce adversarial perturbations that induce changes in human labeling: this means benchmarking does not return a fraction of ``gold-standard'' human performance and instead poses the robustness problem as one that is unsolvable by networks with human-level robustness. Supporting this unsolvability on CIFAR10, the human limits we find (Figure \ref{fig:teaser}'s red line) match our scaling laws' asymptotes (e.g., the FID=0 limit is the top of the shaded green area in Figure \ref{fig:teaser}).

%\section{Estimating the scaling behavior of adversarial robustness}
\section{Scaling Laws for Adversarial Robustness}
\label{sec:scaling_laws}

\iffalse{
\begin{tcolorbox}[enhanced,
  colframe=blue!70!black, colback=blue!10!white, coltitle=white,
  fonttitle=\bfseries, boxrule=0.5mm,
  sharp corners, rounded corners=southeast,
  drop shadow=black!50!white, shadow={1mm}{-1mm}{0mm}{black!20!white},
  title=Motivating Question]

  \textbf{Q:} Can state-of-the-art techniques solve a simple robustness benchmark problem when their available computational resources are scaled?

  \textbf{A:} No, performance asymptotes well before 100\%. 

\end{tcolorbox}
}
\fi

In this section, we discuss the methodology for our scaling laws. We start by training a range of models, varying model size, dataset size, and synthetic dataset quality. We then use these models' test data performances to fit empirical estimators of how the loss should vary with these factors. 
%and agree on the slow scaling and limits shown in Figure \ref{fig:teaser}, which was made using our Approach 2. 

%It is possible that some seemingly simple robustness benchmark problems can be solved by efficiently allocating a sufficient amount of training resources to state-of-the-art adversarial training techniques. To test this, we train various models using established methods, then fit scaling laws to the resulting performances. Using settings predicted to be compute optimal by these scaling laws, we can efficiently match and surpass prior robustness milestones. However, these same scaling laws also suggest a peak below 100\%, consistent with human performance limits stemming from misalignment of human and data labeling (see Section \ref{sec:human}).

\subsection{Background and Methodology}

\paragraph{Adversarial Robustness} Both adversarial attacks and defenses can be understood using the following formulation:
\begin{equation}
\label{eq:robustness_problem}
\small \minimize_{\theta} \rho(\theta),\mathrm{where}~ \rho(\theta) = \mathbb{E}_{(x,y) \sim \mathcal{D}} \left[ \maximize_{\delta \in \mathcal{S}} L(f_\theta( x + \delta), y) \right]
\end{equation}
\noindent and here $\mathcal{D}$ is the data distribution over image-label pairs $(x,y)$, $f$ is the neural network parameterized by $\theta$, $L$ is the cross-entropy loss, and the set of allowable perturbation vectors $\delta$ that may be applied to $x$ is $\mathcal{S} = \{ \delta ~|~ \| \delta \|_{\infty} \leq 8/255 \}$. The adversarially attacked image $x+\delta$ created in the maximization step of Equation \ref{eq:robustness_problem} can be found via $t=1$ \citep{goodfellow2014explaining} or $t>1$ \citep{madry2017towards} steps of projected gradient descent (PGD) on the negative loss: 
\begin{equation}
\label{eq:pgd}
\begin{aligned}
&x^{t}=x+\delta^{t}, \mathrm{where}
\\
& \delta^{t} = \proj_{\mathcal{S}} \left( \delta^{t-1} + \alpha \sign (\nabla_{\delta^{t-1}} L(f_{\theta}(x^{t-1}), y))\right), 
\end{aligned}
\end{equation}
$\delta^{0}$ is a random initial perturbation in the set $\mathcal{S}$, and $\proj_{\mathcal{S}}(\delta)$ projects the candidate perturbation $\delta$ onto the set $\mathcal{S}$ so that no pixel changes by more than $8/255$. Defenses against such attacks can be improved by attacking data at training time, i.e., by optimizing $\rho(\theta)$ or ``adversarially training''. Performances on adversarial and regular datasets can be balanced by using the TRADES loss \citep{zhang2019theoretically}. 

\paragraph{Adversarial Training Settings} We follow the adversarial training approach of \citet{wang2023better}, a leading CIFAR10 approach on the RobustBench leaderboard \citep{croce2020robustbench}. Specifically, we use: 10 PGD steps to adversarially attack images at training time with step size $\alpha=2/255$; label smoothing \citep{szegedy2016rethinking} $0.1$;
synthetic CIFAR10 data from generative models in our training datasets; the TRADES loss \citep{zhang2019theoretically} with $\beta=5$; weight averaging \citep{izmailov2018averaging} with decay rate $\tau= 0.995$; SGD optimization with Nesterov momentum \citep{nesterov1983method} set to $0.9$ and weight decay $5\times10^{-4}$; and a cyclic learning
rate schedule with cosine annealing \citep{smith2017super}. The learning rate and batch size we use varies based on dataset size according to optimal settings found by hyperparameter search: datasets with 10M or fewer samples use batch size 1024, otherwise batch size is 2048; the learning rate is 0.3 for datasets with 10M or fewer samples, 0.2 for larger datasets with up to 200M samples, and 0.1 for datasets with 300M samples (see Appendix \ref{sec:hparam}).

\begin{figure*}[t]
    \centering
    \includegraphics[width=1\linewidth]{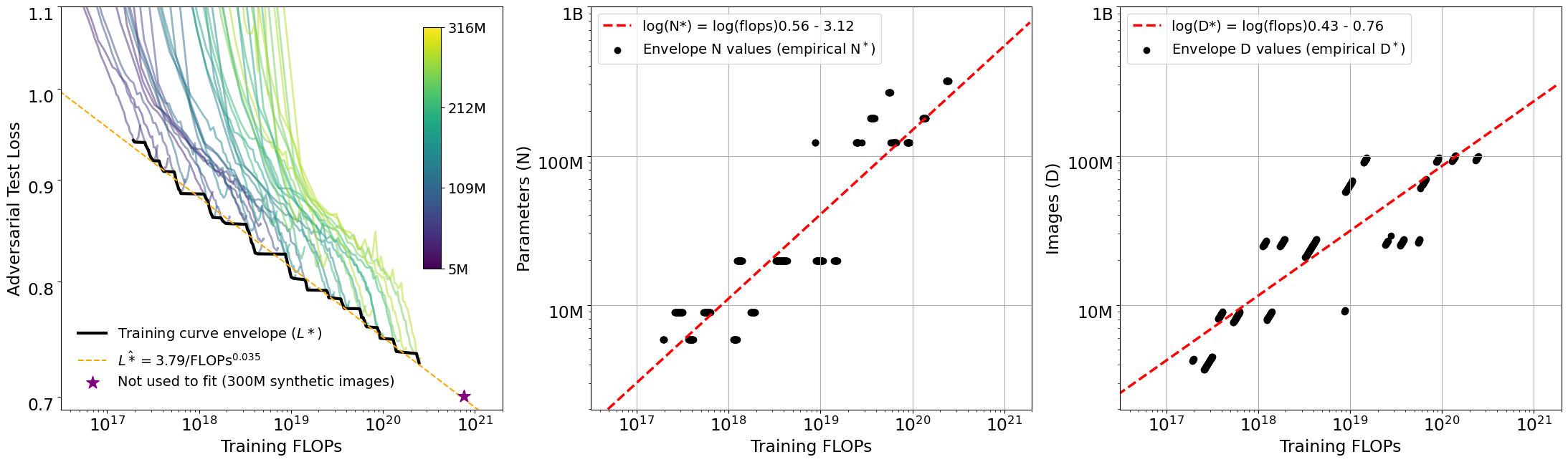}
    %\includegraphics[width=1\linewidth]{figs_and_tables/Figure_2_w_alpha.png}
    %\vspace*{-8mm}
    \caption{\textbf{Approach 1 fits to the envelope of training curves.} This adversarial robustness scaling law approach does not model the effect of training dataset FID on performance, so it is only fit to results associated with one training dataset. For example, in the \textbf{left} plot, we show the learning curves of our models trained on PFGM++ synthetic data. When training, we vary $N$ between 6M and 316M and $D$ between 5M and 100M. Given the optimal $N,D$ combination at each FLOP level (the ``envelope''), we estimate how the optimal $N$, $N^*$ (\textbf{middle}), and optimal $D$, $D^*$ (\textbf{right}), changes as a function of FLOPs via log-space regressions on points sampled from the envelope.}
    \label{fig:approach1}
\end{figure*}

\paragraph{Training Datasets} We build training datasets from three different synthetic data generators. When training models used to fit scaling laws, datasets contain synthetic CIFAR10 data of one specific quality (FID), and we generate enough data to only see each example once during training, allowing our scaling laws to reflect the benefit of adding a unique data point. Note that this mirrors how scaling laws in the LLM literature are learned; e.g. \citet{hoffmann2022training} train on each data source for one epoch, with some minor exceptions (e.g., Wikipedia is seen $\sim$$3\times$). Our approach has similar, rare exceptions (see Appendix \ref{sec:data}).

%\textcolor{red}{
%\paragraph{Training Datasets}
% We build training datasets from three different synthetic data generators in two different ways:  
% (1) when training models with SOTA robustness or efficiency, we mix CIFAR10 data and synthetic data \citep{wang2023better} at a rate of 1 CIFAR10 example for every 7 synthetic examples to boost accuracy and robustness. 
% (2) When training models for scaling laws, we follow the LLM community's methodology to build such laws~\citet{hoffmann2022training}. During training we generate and only use synthetic CIFAR 10 data to see each example only once\footnote{LLM scaling laws in \citet{hoffmann2022training} use Wikipedia data $\sim3\times$. We have similar rare exceptions.} capturing the benefits of adding a unique data point for one epoch. For further information see (see Appendix \ref{sec:data}).
%}

We generate synthetic data via the elucidating diffusion model (EDM) \citep{karras2022elucidating}, the Poisson Flow Generative Model plus plus (PFGM++) \citep{xu2023pfgm++}, and Discriminator Guidance (DG) \citep{kim2022refining}. We use each generator and a variety of diffusion steps to create 7 training dataset generators with various image qualities  (see Table \ref{tab:datasets}), which we measure with FID \citep{heusel2017gans}. 

\begin{table}
\caption{Our synthetic data generators and their image qualities.}
\label{tab:datasets}
\vskip 0.15in
\centering
\begin{widetabular}{\linewidth}{lccccccc}
\toprule
& \multicolumn{7}{c}{\textbf{Synthetic Data Generators}} \\ 
Type: & EDM & EDM & EDM & EDM & EDM & PFGM++ & DG \\ 
Steps: & 5 & 6 & 7 & 10 & 20 & 18 & 20 \\ 
\midrule
FID     & 35.54  & 14.26   &  6.79  &  2.48  & 1.82  & 1.76 &  1.65 \\ 
\bottomrule
\end{widetabular}
\end{table}

For each training dataset generator, we generate 100 million (EDM-7, EDM-20, PFGM++, DG) or 30 million (EDM-\{5,\;6,\;10\}) unique samples. For generators with 100 million samples, we train models on 5, 10, 30, 70, and 100 million samples, corresponding to traditional CIFAR10 epoch counts of 100, 200, 600, 1400, and 2000. The other generators are only used to train on 5,10, and 30 million samples.

\paragraph{Models} We train the WideResNet models \citep{zagoruyko2016wide} shown in Table \ref{tab:models}. \citet{peng2023robust} boost the prior SOTA AutoAttack performance of \citet{wang2023better}, $70.69$\% to $71.07$\%, by modifying their WideResNet architecture, but we opt for WideResNet as it is a widely-used architecture for this problem \citep{croce2020robustbench}. All models use SiLU activations \citep{hendrycks2016gaussian}.

\begin{table}
\caption{Each model we use and its parameter count $N$ in millions.}
\label{tab:models}
\vskip 0.15in
\centering
\begin{widetabular}{\linewidth}{lcccccccc}
\toprule
& \multicolumn{8}{c}{\textbf{WideResNet Configurations}} \\ 
Depth: & 28 & 40 & 82 & 28 & 58 & 82 & 70 & 82\\  %& 94
Width: & 4 & 4 & 4 & 12 & 12 & 12 & 16 & 16\\  %& 16 
\midrule
$N\times 10^{-6}$     & 6 &  9 &  20 &  53 & 122 & 178 & 267 & 316 \\  %& 366
\bottomrule
\end{widetabular}
\end{table}

\paragraph{Evaluation} In Section \ref{sec:efficient}, we report accuracy on AutoAttack \citep{croce2020reliable}, which performs 4 unique attacks (including modified versions of PGD) and is tracked by the RobustBench leaderboard \citep{croce2020robustbench}. In our scaling law plots, we report either model loss or accuracy, which are computed on the CIFAR10 test set after its images were adversarially attacked using 40 PGD steps on the model's CW loss \citep{carlini2016towards}. Note that the reported adversarial loss is the TRADES loss with $\beta=5$, mixing vanilla and adversarial performances. Also note that the adversarial accuracies that result from this PGD attack tend to be about 1\% higher than those resulting from AutoAttack  \citep{wang2023better} but are significantly cheaper to obtain. We derive FLOPs values via the approach in Appendix \ref{sec:FLOPs}.
%In Figure \ref{fig:teaser}, estimated optimal accuracies are derived via a regression of this adversarial accuracy on loss because our scaling laws use losses, creating a need for translation to accuracy (Appendix \ref{sec:loss_acc} shows the good fit of this regression).

%To understand the efficiency of SOTA adversarial training techniques on a common robustness benchmark problem, we derive scaling laws on CIFAR10 \citep{krizhevsky2009learning} models trained to be robust to adversarial pixel perturbations of maximum size $\ell_{\infty} = 8/255$. The performance on this problem is tracked by RobustBench \citep{croce2020robustbench}, which lists 71.07\% as the current state of the art AutoAttack accuracy. 

\subsection{Deriving Scaling Laws}
In this section, we explore three parametric forms for our scaling laws to fit the data generated from our large-scale experiments. Next, we use our scaling laws to recommend: a) the optimal resource allocation (i.e., model and dataset size) at various training FLOPs values, allowing robustness to be maximized given a budget; and b) the optimal return on investment, i.e., the best robust loss/accuracy that can be achieved for a given compute budget.
Finally, we also derive the tradeoff between model size and training compute overhead, showing that there are opportunities to reduce model size (and improve inference speed).

%We find scaling laws using three approaches inspired by the LLM literature \citep{kaplan2020scaling,hoffmann2022training}.

\subsubsection{Approach 1: Envelope of training curves}

Our first approach to understanding the scaling behavior of adversarial training is inspired by the ``Approach 1'' used in the Chinchilla scaling study \citep{hoffmann2022training}. Specifically, given a collection of 8 models with parameter counts $N$, each trained on up to 5 different dataset sizes $D$, we find the lowest loss at each training FLOP value -- these are the points on the black envelope line in Figure \ref{fig:approach1}. We fit a power law to predict the lowest loss $L^*$ at these compute budgets. We also identify the $N,D$ combination with the lowest loss at each training FLOP value. Using these optimal configurations, we fit power laws to predict the optimal $N,D$ from the training FLOPs budget, obtaining relations of the form  $N^* \propto \mathrm{FLOPs}^a$ and  $D^* \propto \mathrm{FLOPs}^b$.

We repeat Approach 1 for each data generator, and we find similar results for data generators with similar qualities (see Appendix \ref{sec:approach_1_app} for more on Approach 1). However, as we improve data quality---e.g., moving from EDM-5 to EDM-6 to EDM-20---we notice two changes: (1) the minimum loss at each FLOP level improves; and (2) the exponents $a,b$ move closer towards $0.5$, the values they took in \citet{hoffmann2022training}. For example, $a=0.73, b=0.27$ for EDM-5, but the PFGM++ results shown in Figure \ref{fig:approach1} have $a=0.56, b=0.43$, which is consistent with the idea that lower quality data is less beneficial \citep{wang2023better}. Moreover, the fact that improving data quality raises $b$ and lowers the loss suggests that a rigorous understanding of scaling adversarial training must account for data quality, consistent with results on non-synthetic, non-adversarial CIFAR10 scaling \citep{sorscher2022beyond} and motivating our Approaches 2 and 3. Indeed, unlike Approach 1, Approaches 2 and 3 can fit to training runs that used various FIDs, and predict performance as a function of FID.

\begin{figure*}[t]
    \centering
    \includegraphics[width=.33\linewidth]{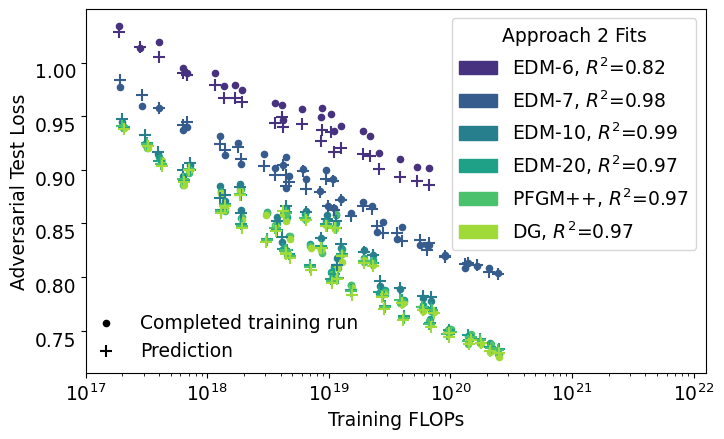}
    \includegraphics[width=.33\linewidth]{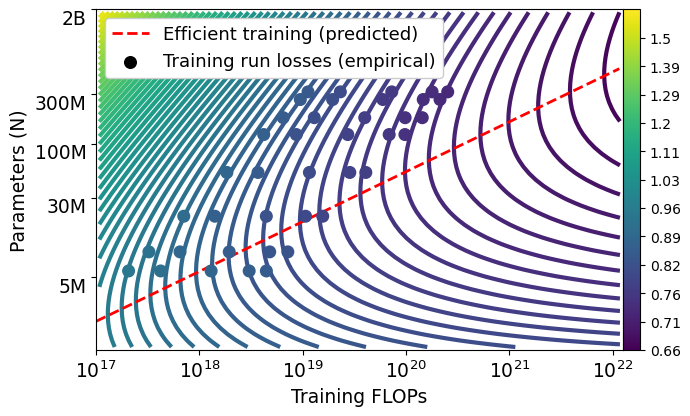}
    \includegraphics[width=.33\linewidth]{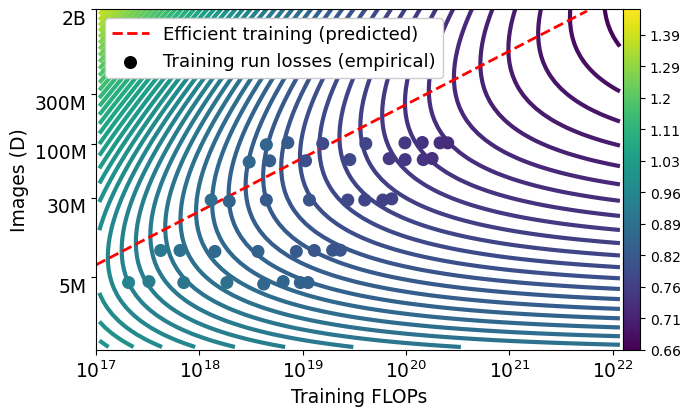}
    \caption{\textbf{Approach 2 models the effect of synthetic dataset quality.} Approach 2 can fit to training runs that used datasets with differing FIDs by modeling FID's effect on performance. (\textbf{left}) Equation \ref{eq:approach_2} fits runs that used different models and dataset sizes/qualities. EDM-6 and PFGM++ were not used to learn the fit's parameters, but are plotted to show extrapolation performance of the fit. We plot isoLoss contours (regions with the same loss predicted by our scaling laws; higher losses are brighter) and the curves for $N^*$ (\textbf{middle}) and $D^*$ (\textbf{right}), the training settings predicted to be compute-optimal at a FLOPs budget for a synthetic dataset with DG20's quality.}
    \label{fig:approach2_fit}
\end{figure*}

\subsubsection{Approach 2: Parametric loss with data quality terms}
\label{sec:approach_2}

Approach 2 addresses the inability of Approach 1 to model two important limiting cases: (1) what happens when $N,D$ tend toward infinity; and (2) what happens when  $N,D$ are at practical levels but the training dataset has unlimited high-quality CIFAR10 examples (i.e., an FID=0 synthetic dataset). For example, as shown in Figure \ref{fig:teaser}, Approach 2 can predict performance on an FID=0 training dataset and the asymptotic performance as a function of FID.\footnote{In Figure \ref{fig:teaser}, Approach 2 predicts optimal losses, then we convert these losses to accuracies %using a regression-based predictor of accuracy 
(see Appendix \ref{sec:loss_acc} for details).}

To achieve this, Approach 2 parameterizes the loss with the functional form used in ``Approach 3'' of \citet{hoffmann2022training}, but it modifies two constants related to the ability of $D$ to affect asymptotic performance so that they are functions of data quality as shown in Equation \ref{eq:approach_2}:
\begin{equation}
\label{eq:approach_2}
\begin{aligned}
    &L(N,D,\mathrm{Quality}) = \frac{A}{N^{\alpha}} + \frac{B'}{D^{\beta}} + E', 
    \\
    &E' = \mathrm{exp}(\mathrm{log}(E) + \mathrm{log}(1+\mathrm{Quality}^{-1})\epsilon), 
    \\
    &B' = \mathrm{exp}(\mathrm{log}(B) + \mathrm{log}(1+\mathrm{Quality}^{-1})\zeta),
\end{aligned}
\end{equation}
where $\epsilon$ and $\zeta$ model how changes in data quality affect the irreducible loss $E$ and the portion of the loss reducible by dataset growth $B$, respectively, and $\mathrm{Quality}=\frac{1}{\mathrm{FID}}$. Note that with infinitely high synthetic data quality (i.e., FID=0 data), Equation \ref{eq:approach_2} simplifies to $L = A/N^{\alpha} + B/D^{\beta} + E$. 

To learn the parameters of Equation \ref{eq:approach_2}, we first fit all but $\epsilon, \zeta$ to 40 training runs on the DG dataset (8 models trained on DG's 5 sizes). We then hold the learned parameters constant and fit $\epsilon, \zeta$ to models trained on three datasets of varying data quality (EDM-7, EDM-10, EDM-20). We use the L-BFGS optimization approach described in \citet{hoffmann2022training} for both fitting procedures (see Appendix \ref{sec:approach_2_app} for more details on the fitting procedure). In Figure \ref{fig:approach2_fit}, we show how the fitted parameters accurately predict loss for the dataset used to fit the base parameters (DG), the datasets used to fit the quality parameters (EDM-7, EDM-10, EDM-20), and two held out datasets (EDM-6, PFGM++). 

Figure \ref{fig:approach2_fit} also shows the optimal $N,D$ for each training FLOPs level on the DG dataset. We derive these curves by analytically minimizing Equation \ref{eq:approach_2} with respect to $N$ ($D$) after replacing $D$ ($N$) using the constraint $\mathrm{FLOPs}\approx7822ND$, rather than the constraint $\mathrm{FLOPs}\approx6ND$ used for non-adversarial LLM training \citep{kaplan2020scaling, hoffmann2022training}. Our constraint is found to never have more than 1\% relative error across all observations of our training runs (more than 9000 observations covering 233 different model-dataset combinations). This minimization gives $N^*$ ($D^*$) as a power law function of FLOPs and the exponent $a$ ($b$): $N^* \propto \mathrm{FLOPs}^a$ ($D^* \propto \mathrm{FLOPs}^b$). Specifically, we have the following:

\footnotesize
\begin{equation}
\label{eq:n_star_d_star}
N^*=G{\left(\frac{\mathrm{FLOPs}}{7822}\right)}^a,\ \
D^*= G^{-1}{\left(\frac{\mathrm{FLOPs}}{7822}\right)}^b,
\end{equation}
\normalsize
where $G = \left(\frac{\alpha A}{\beta B}\right)^{\frac{1}{\beta+\alpha}}, a={\frac{\beta}{\beta+\alpha}}, b={\frac{\alpha}{\beta+\alpha}}$.

\subsubsection{Approach 3: Parametric loss modeling data quality and its effect on overfitting}
\label{sec:approach_3}

While Approach 2 models the effect of synthetic data quality, it is possible that its functional form does not capture the complex relationship between data quality and loss scaling. Indeed, Approach 2 models quality rather simply, and empirically
%we find that it struggles to simultaneously fit all of our datasets when we include EDM-5 (FID=35.54). 
we find that the fit of this simple power law form is poor when the fitting includes the lowest quality dataset (EDM-5, FID=35.54). 
Accordingly, we test the validity of the scaling laws derived by Approach 2 by using a more complex functional form that can simultaneously model the loss on all datasets accurately.

Our Approach 3 is inspired by Equation 4.1 of \citet{kaplan2020scaling}, which models how increasing $N$ causes overfitting (and loss plateauing) earlier as $D$ is reduced. Here, rather than modeling this ``data size bottleneck'' that restricts the benefit of increasing $N$, we model a ``data quality bottleneck'' that restricts the benefit of increasing $D$. Compared to \citet{kaplan2020scaling}, we use a similar functional form but couple $D$ with $\mathrm{Quality}$ rather than with $N$: 
\begin{equation}
\scriptsize{
\begin{aligned}
\label{eq:approach_3}
&L(N,D,\mathrm{Quality}) = \frac{A}{N^{\alpha}} + \left(\frac{B}{D} + \left(\frac{Q}{\mathrm{Quality}}\right)^{(\kappa/\beta)}\right)^{\beta} + E',
 \\
&\mathrm{where}\ \ E' = E + \mathrm{log}(1+\mathrm{Quality}^{-1})\epsilon.
\end{aligned}
}
\end{equation}

\begin{figure}
    \centering
    \includegraphics[width=1\linewidth]{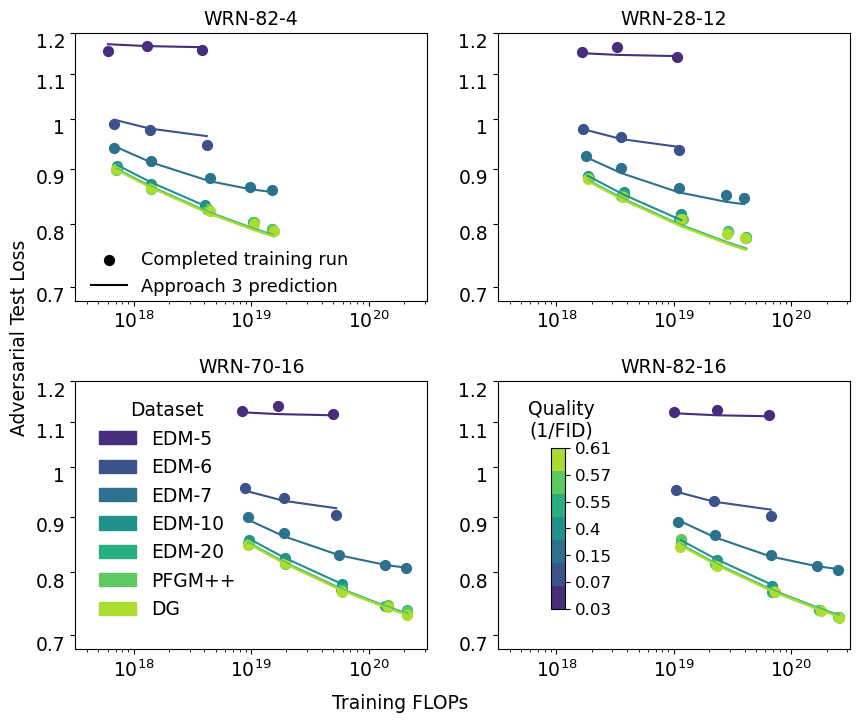}
    %\vspace*{-5mm}
    \caption{\textbf{Approach 3 fits to models trained on data of any FID.} Our third approach can account for overfitting happening sooner on lower quality datasets, allowing it to simultaneously model training runs from all of our datasets, even those with very low qualities. As a result, Approach 3's learned parameters account for more training run results than other approaches' parameters do.}
    \label{fig:approach3_fit_4_models}
\end{figure}

Like Approach 2, our Approach 3 simplifies at FID=0 to $L = A/N^{\alpha} + B/D^{\beta} + E$, assuming that the constant $B$ is set to $B^{\beta}$. We fit Equation \ref{eq:approach_3} using all datasets and models, finding $L$ is accurately predicted due to the ability to model overfitting as a function of data quality (see Figure \ref{fig:approach3_fit_4_models}). Since Equation \ref{eq:approach_3} lacks a simple derivative with respect to $N,D$ at nonzero FIDs, we numerically solve for its $N^*,D^*$. Appendix \ref{sec:approach_3_app} further discusses Approach 3 and its L-BFGS fitting, and Figure \ref{fig:approach3_fit} extends Figure \ref{fig:approach3_fit_4_models} to show all 8 models.

Ultimately, we obtain similar results with each approach (Table \ref{tab:approach_compare} compares them), bolstering support for our scaling-law-related conclusions. Also, Figure \ref{fig:approach_compare} shows that Approaches 2 and 3 have nearly identical $N^*$ predictions for $\mathrm{FID}=0$ (and shows all approaches' $N^*$ for other FIDs).

\begin{table}[t]
\caption{\textbf{All scaling law approaches suggest similar conclusions.} Comparison of $a$, $b$, $L^*$ at $10^{25}$ FLOPs, and $E$ (if modeled).}
\label{tab:approach_compare}
\vskip 0.15in
\centering
\begin{widetabular}{1.07\linewidth}{lcccccccccc}
\toprule
& \multicolumn{4}{c}{PFGM++ (FID=1.76)} & \multicolumn{4}{c}{Hypothetical (FID=0)} \\ 
 \cmidrule(lr){2-5} \cmidrule(lr){6-9} 
Approach  & $a$ & $b$ & $L^*$ & $E'$  & $a$ & $b$ & $L^*$ & $E$  \\ 
\midrule
1 & 0.56 & 0.43 & 0.50 &  --  &  --  &  --  &  --  &  -- \\
2 & 0.48 & 0.52 & 0.61 & 0.56  & 0.48 & 0.52 & 0.53 & 0.48 \\
3 & -- & -- & 0.63 & 0.55  & 0.47 & 0.53 & 0.56 & 0.52 \\
\bottomrule
\end{widetabular}
\end{table}

\subsection{Putting Scaling Laws into Practice}
\label{sec:efficient}
\subsubsection{Compute-Efficient Adversarial Training}

We use our scaling laws to predict compute-optimal training settings ($N^*, D^*$) at a variety of compute budgets. 
%We train models using these settings at FLOPs values below, matching, and above the prior SOTA's training FLOPs (see Figures \ref{fig:all_n_stars} and \ref{fig:overhead}).
As shown in Figure \ref{fig:all_n_stars}, we train models according to these settings and find AutoAttack accuracy surpasses the prior SOTA's \citep{wang2023better}. 
Thus, past inability to reach higher robustness levels is partially addressed via more efficient training settings.
Further scaling also helps robustness.

We highlight the following results (see Figure \ref{fig:all_n_stars}):
\begin{enumerate}[label={\bfseries (\arabic*)}]%
%\begin{enumerate}
    \item A new SOTA of $73.71$\% AutoAttack ($93.68$\% clean) accuracy by training WideResNet-94-16 on 500M samples.
        
    \item A model predicted to be compute-optimal at the prior SOTA's training FLOPs budget (WideResNet-82-8, 79M parameters) beats the prior SOTA (WideResNet-70-16, 267M parameters) by 1\% AutoAttack, reaching $71.59$\% AutoAttack and $93.11$\% clean accuracy, while being over $3\times$ more FLOP and parameter efficient.
    
    \item A model trained entirely on synthetic data outperforms the prior SOTA, attaining $71.7$\% AutoAttack accuracy and $93.27$\% clean accuracy.
    
\end{enumerate}

Given the benefit of training on higher quality data observed in all of our approaches (e.g., Figure \ref{fig:teaser}), we boost dataset quality for these training runs by mixing CIFAR10 data in with the synthetic data at a rate of 1 CIFAR10 example for every 7 synthetic examples. Such mixing was also done to obtain the prior SOTA \citep{wang2023better}. However, note that we also train one model entirely on synthetic data.

%More details %on these results 
%are available in Appendix \ref{efficient_app}. 
In Appendix \ref{efficient_app}, we explore Figure \ref{fig:all_n_stars} and Table \ref{tab:approach_compare} more thoroughly. For example, we discuss the variation in $N^*$ predictions and why Approach 3 provides a bridge between the other two approaches. 
Of note, while 
$N^*$ predictions vary by approach,
%there is variation in our approaches' $N^*$ predictions, 
all of our scaling laws agree that the prior SOTA was trained with too large a model, which is just outside the compute-efficient region in Figure \ref{fig:all_n_stars}.

\subsubsection{Trading off Training Efficiency for Inference Efficiency}

In real-world scenarios, it is often helpful to deviate from compute-optimal settings by using extra resources to train a smaller model longer, gaining memory and inference efficiency without harming performance. To guide this decision making process, we study the tradeoff between model size and compute overhead \citep{kaplan2020scaling,overhead}. 

Given a desired model size that is $\omega_N\times$ as large as the compute-optimal model size $N^*$, we can find the multiple $\omega_D$ of the optimal dataset size $D^*$ needed to maintain the expected loss of the optimal model and dataset sizes. In particular, we ensure no change in our Approach 2 scaling law predicted loss by choosing $\omega_D$ as follows:

{
\small
\begin{equation}
\begin{aligned}
\label{eq:overhead}
&E' + \frac{A}{{N^*}^{\alpha}} + \frac{B'}{{D^*}^{\beta}} = E' + \frac{A}{\left(\omega_N {N^*}\right)^{\alpha}} + \frac{B'}{\left(\omega_D {D^*}\right)^{\beta}},
 \\
&\mathrm{where}\ \ \omega_D = \left(1 - (\omega_N^{-\alpha} - 1) \frac{A {N^*}^{-\alpha}}{B' {D^*}^{-\beta}}\right)^{-\frac{1}{\beta}}
\end{aligned}
\end{equation}
}

and $E'$ and $B'$ assume a particular dataset quality. Given new data and model sizes, the ``compute overhead'' is the increase in training FLOPs as a percentage of the training FLOPs needed for the optimal model and dataset sizes.

\begin{figure}[t]
    \centering
    \includegraphics[width=1\linewidth]{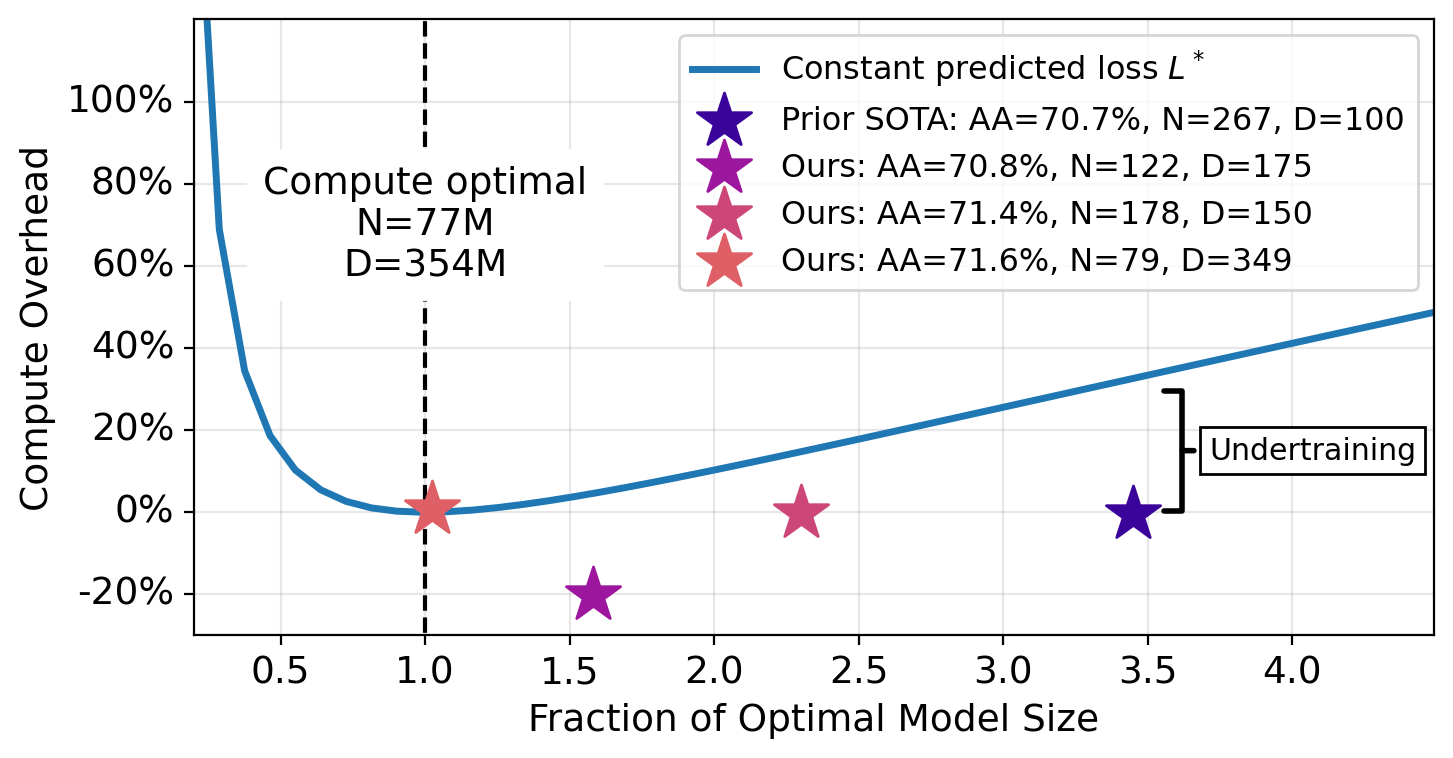}
    %\caption{\textbf{Many model sizes achieve the same loss.} Using the FLOPs budget of the prior SOTA, we can train a variety of model sizes to the same expected loss. Our scaling laws suggest existing approaches train models that are too large, and larger datasets are needed to near compute-optimal training.}
    \caption{\textbf{Train smaller models on larger datasets.} Given the FLOPs budget of the prior SOTA, we find the optimal model size using Approach 2 (assuming DG-20 data), its predicted loss $L^*$, and the expected compute overhead incurred by training other model sizes to  $L^*$. The prior SOTA is not expected to reach $L^*$ due to its small training dataset---consistent with this, we train smaller models on more data and find AutoAttack accuracy increases.}
    \label{fig:overhead}
\end{figure}

As can be seen in Figure~\ref{fig:overhead}, there exists a substantial region where the model size can be reduced with minimal compute overhead. For example, reducing the model size by half ($\omega_N=0.5$) results in only a $\sim$$10$\% increase in computational requirements, demonstrating an efficient tradeoff between model compactness and training cost. We also show that, given the prior SOTA's training FLOPs (fixing compute overhead to 0\%), larger models are predicted to require increasingly more compute to reach the performance of the compute-optimal model, consistent with the displayed AutoAttack performances falling as model size increases.

%Here, we test whether our scaling laws can be used to find $N,D$ that leads to robustness that matches the prior SOTA at a smaller cost (more efficiency) and robustness that surpasses the prior SOTA if allowed a larger budget (higher robustness). To achieve this, we select two training FLOPs budgets, one 20\% lower than the prior SOTA's and one \textcolor{red}{XX\% higher than the prior SOTA's}, then we choose model sizes within the range suggested to be optimal by our different approaches. 

%we use our scaling laws to find compute-efficient settings of training parameters (see Figure \ref{fig:all_n_stars}). 

%our scaling laws suggest there are limits to the CIFAR10 robustness that is attainable

%if there are limits, that is bad for foundation models?

%if attacks during benchmarking created invalid data, that would support the existance of limits to adversarial robustness that are implied by the asymptotic limits of our scaling laws. 

\section{Small-Scale Human Performance Study}
\label{sec:human}

\begin{figure}[t]
    \centering
    \includegraphics[width=1\linewidth]{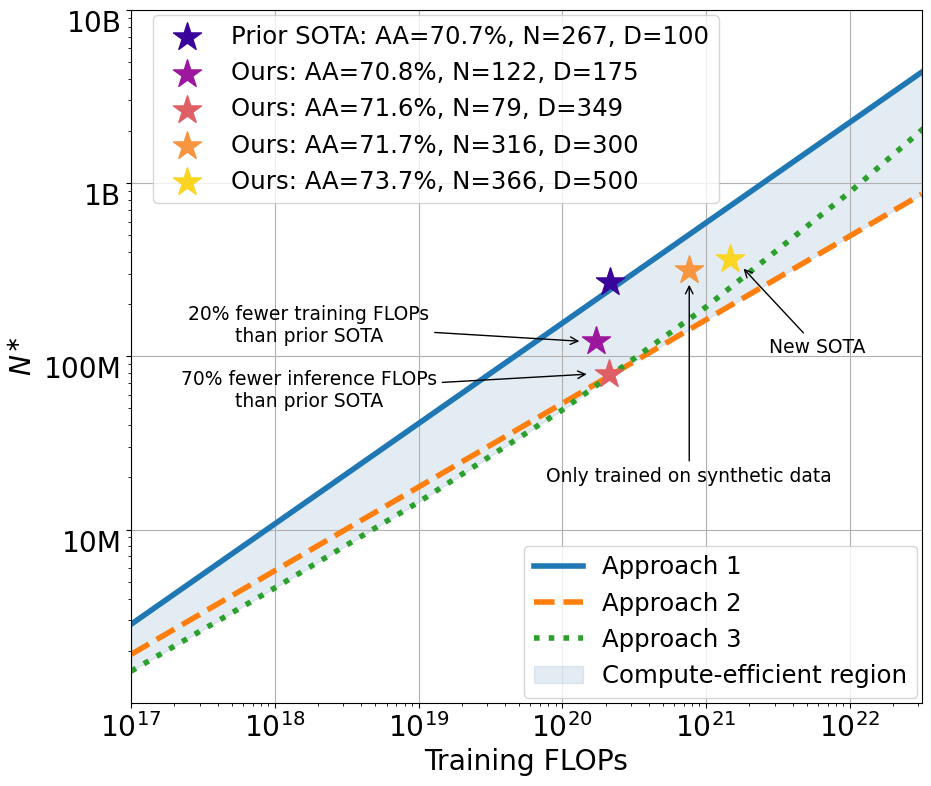}
    %\vspace*{-5mm}
    \caption{\textbf{Training settings estimated to be compute-efficient by our scaling laws are state-of-the-art, validating our scaling law fits.} We find $N^*$ for the DG-20 dataset for Approaches 1--3 and train models using values of $N$ in between at least two approaches' predicted values for $N^*$ (see the ``compute-efficient region''). The trained models are more efficient and robust than the prior SOTA.}
    \label{fig:all_n_stars}
\end{figure}

The irreducible error captured by our scaling laws (Table \ref{tab:approach_compare}) indicates that, even with unlimited compute, the CIFAR10 $\ell_{\infty}$ AutoAttack benchmark problem cannot be solved. % (i.e., benchmark accuracy plateaus around $91\%$).
Here, we investigate whether this irreducible error is attributable to a flaw in our scaling laws or a fundamental intractability of achieving robustness to $\ell_p$-norm-bounded attacks.

%In particular, one possibility is that additional data would lead to more accurate scaling laws that have 0 irreducible error.
%This performance limit suggests that our scaling laws may be overly pessimistic:
One possibility, supported by recent work \citep{gaziv2023robustified}, is that adversarial attacks on highly robust models produce adversarial images containing no clear object or an object of the wrong class, which we call \textit{invalid adversarial data}. If attacks during benchmarking create such invalid data, then irreducible error (or a limit to benchmark performance that is below $100$\%) would be expected because both humans and neural networks would by definition fail to classify such images correctly (except by chance). 
%not appropriately evaluating adversarial alignment.
%In fact, recent work by \citet{gaziv2023robustified} has suggested that this may in fact be the case.

Hence, in an effort to understand the cause of the irreducible adversarial error in our scaling laws, we perform a small-scale human performance evaluation on the 2629 adversarial images that were incorrectly classified by our SOTA model (with $73.71\%$ AutoAttack accuracy).
%At a high level, can we align AI's response on adversarial images with human response on them.
%In particular, we use human performance to understand the extent to which our SOTA NN's performance is limited by invalid data.
%shed light on why the SOTA neural network fails on these images and to determine if any images produced by adversarial attacks with our SOTA model are no longer aligned with the human labels (i.e., ground truth).
%We posit that \textcolor{red}{adversarial misalignment is the underlying cause of nearly all of the irreducible adversarial error found in our scaling law (rethink/clarify)} and, further, that it carries implications for benchmarking and training.
Through this study, we
\begin{enumerate*}[label=\textbf{(\arabic*)}]%
    \item \emph{Identify that invalid adversarial data is used when benchmarking CIFAR10 robustness with $\|\delta \|_{\infty}=8/255$ attacks}: 
    %of the $2629$ adversarial images that cannot be classified by our SOTA network, 
    Human performance suggests that $727/2629$ images are ``invalid''. This finding characterizes the extent to which {existing benchmarking does not track the fraction of gold-standard ``human performance'' achieved by the NN}. In other words, it suggests the need for an improved attack formulation that takes data validity into account.
    
    \item \emph{Propose a revised benchmarking scheme} to address the nonzero irreducible error of the $l_{\infty}$ attack formulation: human-identification of invalid images due to class deception or ambiguity, %resolution,
    then removing invalid data from benchmarking. We also encourage more research in designing attack formulations that take image validity into account. 
    %\item \textbf{Uncover the presence of invalid images during training} after finding that weaker attacks used at train time on our SOTA adversarially robust model can produce training images that humans find inconsistent with their labels.
\end{enumerate*}
%identify subsets of valid and invalid images where invalid images are those that when attacked with a SOTA robust model the adversarial image cannot be confidently correctly classified by either humans or a SOTA multi-modality foundation model, GPT-4.
%Our final categorization suggests that a subset of the CIFAR10 test images should be removed from $\ell_{\infty}$ adversarial benchmark as they are a source of irreducible error even in human evaluation.
%Our key takeaways from this study are:

\subsection{Evaluating Human Performance} \label{sec:misalign-study}
%Here we present analysis and findings from our human and GPT-4 evaluation of the adversarial images produced by running AutoAttack with our SOTA model.
%We included GPT-4 as it represents a more human-like vision classification system that can be notified via prompting that images were distorted.

To understand human performance on benchmarking data, we first saved the adversarial and clean (pre-attack) versions of the $2629$ CIFAR10 test images that were classified incorrectly when running AutoAttack \citep{croce2020reliable} on our SOTA model.
We performed classification of the adversarial images using (1) three human users and (2) GPT-4 (zero-shot classification).
Each user was instructed to classify each image and provide a confidence value of either low or high.
Following classification of the adversarial images, the human users also classified and provided confidence ratings for the clean images. The three human users are all members of our research team. We chose GPT-4 as a fourth user to supplement our small-scale human study, as GPT-4 has strong object recognition abilities and can understand the task via instructions similar to those that we give humans \citep{yang2023dawn}. Additional experimental details including the GPT-4 prompt are available in Appendix \ref{sec:invalid_app}.

\iffalse
\begin{table}
\caption{Classification accuracy of human users and GPT-4 on the 2629 adversarial images incorrectly classified by our SOTA model.}
\label{tab:evaluation-results}
\vskip 0.15in
\centering
\begin{widetabular}{1\linewidth}{lcccc}
\toprule
& \multicolumn{4}{c}{\textbf{Classification Accuracy} (\%)} \\ 
& \multicolumn{2}{c}{Any Confidence} & \multicolumn{2}{c}{High Confidence} \\  
\cmidrule(lr){2-3} \cmidrule(lr){4-5} 
User & Benign & Adversarial & Benign  & Adversarial \\ 
\midrule
Human 1 & 85.17 & 59.76 & 89.05 & 80.32 \\
Human 2 & 92.39 & 72.5 & 98.17 & 98.17 \\
Human 3 & 79.23 & 58.88 & 92.77 & 89.56 \\
GPT-4 & - & 44.12 & - & 44.94 \\
\bottomrule
\end{widetabular}
\end{table}
\fi

\begin{figure}
    \centering
    \includegraphics[width=1\linewidth]{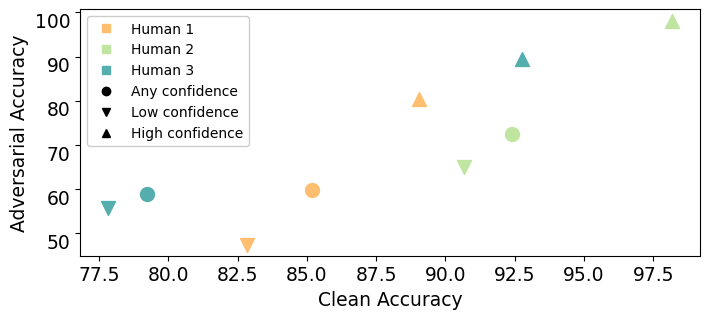}
    \caption{\textbf{Adversarial accuracy vs. clean accuracy by human and confidence.} 
    Classification accuracy of each human user on the clean and adversarial versions of the 2629 images incorrectly classified by our SOTA NN.
    Accuracy is provided for collections of predictions where users indicated any, low, or high confidence.}
    \label{fig:adv-clean-by-human-conf}
\end{figure}

The human user performance on the clean and adversarial versions of the $2629$ images misclassified by our SOTA NN is shown in Figure~\ref{fig:adv-clean-by-human-conf}, showing that average attacked-data classification performance is 25\% lower than average performance on the corresponding clean images. Indeed, every user’s performance drops by at least 21\% when they move from clean to attacked data, supporting the existence of a limit on robustness below 100\% and thus also supporting our scaling laws' modeling of irreducible error on this problem.
%was $85.6\pm5.38$ and $63.71\pm6.22$, respectively. 
%The best human user performance on the adversarial images was $72.5\%$. 
%GPT-4 was only used on the adversarial images and yielded an accuracy of $44.12\%$.
%Analysis of the correctness and confidence of human users 
%and GPT-4 highlighted in Figure~\ref{fig:correct-vs-conf} reveals 
%revealed that high human confidence is a reasonable predictor of correct classification.
%Specifically, $85.6\%$ of the high confidence predictions on adversarial images made by humans were correctly classified.
%However, prediction confidence provided by GPT-4 was a poor indicator of correct classification. 
%Namely, only $44.09\%$ of high confidence classifications made by GPT-4 were correct and $98.1\%$ of all predictions were indicated by GPT-4 to be high confidence. %54.01. 
%However, GPT-4 indicated high confidence $98.07\%$ of the time, yielding incorrect predictions with high confidence on $54.19\%$ of the images. 
%\textcolor{red}{(Add figure with correctness/confidence matrix for human users and GPT-4 to illustrate that high human confidence is a reliable predictor of correct classification but GPT-4 confidence not meaningful)}

Moreover, assuming the human users and GPT-4 were able to correctly classify the $7371$ adversarial images that our SOTA model classified correctly, then the average human accuracy would be $90.46$\% on all of the adversarial images -- for context, GPT-4 would have an accuracy of $85.31\%$ as it obtained $44.12$\% accuracy on the $2629$ adversarial images.\footnote{These numbers do not represent how users would perform if they were adversarially attacked, but rather indicate the extent to which the correct classes are visible in the images that were designed to (and successfully did) attack our SOTA model.}
Interestingly, the limits indicated by our scaling laws and the limits indicated by our human study are consistent, as the average human user accuracy of $90.46$\% closely aligns with the asymptotic adversarial CIFAR10 test accuracy for FID=0 predicted by our scaling law (we illustrate the closeness of these estimated performances in Figure~\ref{fig:teaser}).
%This estimate for the achievable human accuracy on this adversarial test set suggests that roughly \textcolor{red}{$10$}\% of the test images are invalid when measuring the adversarial robustness of a model.
%In other words, these adversarial attacks result in a benchmark producing samples that are misaligned with human perception.
%In other words, for our SOTA robust model, the measurement this adversarial benchmark provides becomes misaligned with achievable human performance on the benchmark.
%can we align AI's response on adversarial images with human response on them.
%Existing benchmark is supposed to measure adversarial alignment with human users but we uncover that this is not the case.
%Existing benchmarks are supposed to measure “adversarial alignment”: how close neural networks are to gold-standard “human performance”.
%However, we perform a human study to reveal that this is not the case.
%Specifically, we identify a subset of images that when adversarially attacked using out SOTA model are consistently incorrectly classified by multiple human users.

\begin{figure}[t]
    \centering
    \includegraphics[width=1.07\linewidth]{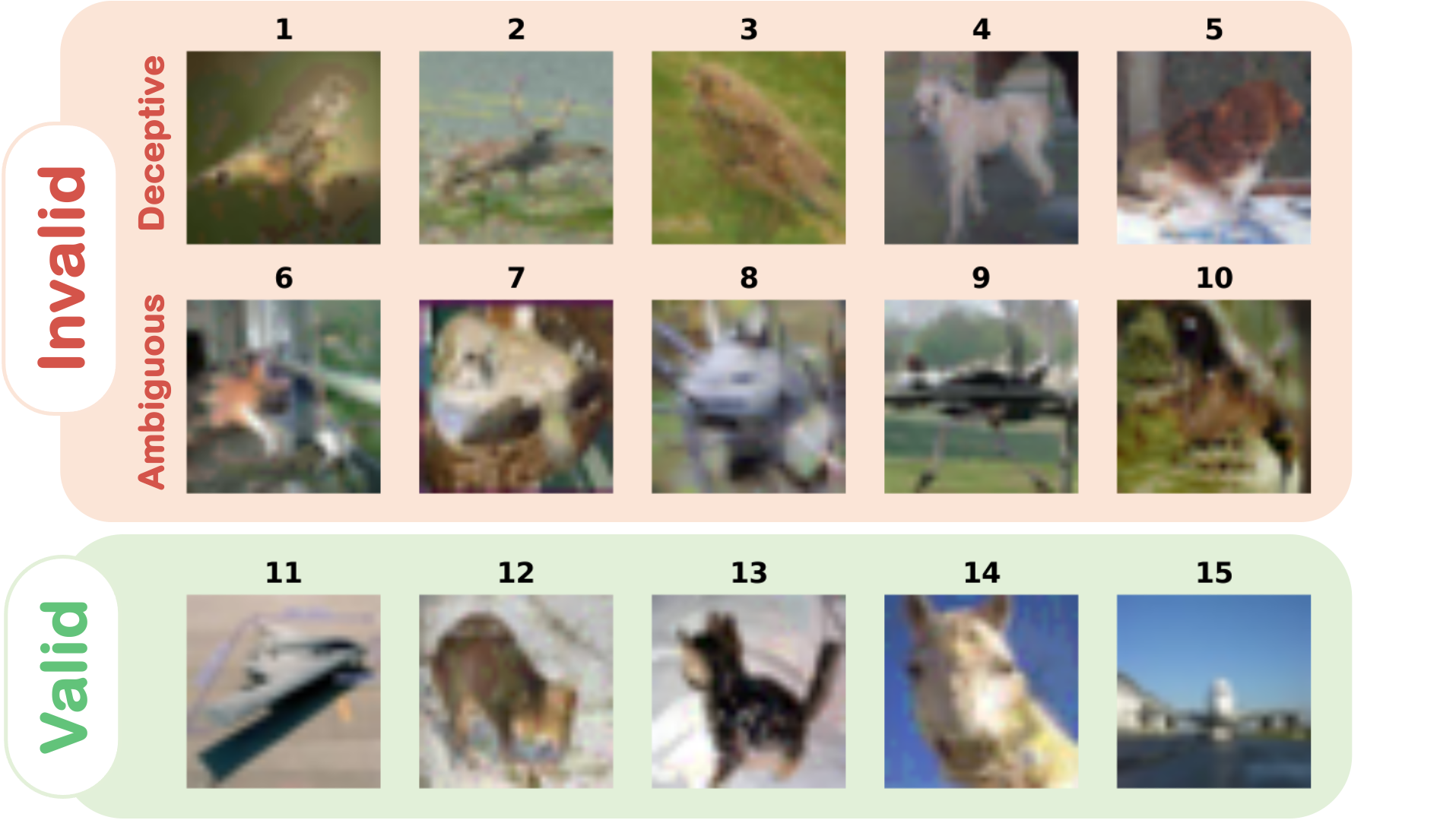} %,trim={10mm 72mm 10mm 0},clip]
    %\vspace*{-6mm}
    \caption{\textbf{A sampling of valid and invalid adversarial images.} 
    %The 2629 adversarial images were categorized as invalid ($27.87\%$ of images) or valid ($72.13\%$ of images) based on analysis of our human and GPT-4 evaluations.
    We categorized 727 out of 2629 adversarial images as invalid based on our human and GPT-4 study.
    %Valid images were those correctly classified by at least half of the users (including GPT-4 as a user) or correctly classified with high confidence by at least one human. %(as GPT-4 confidence was unreliable).
    %All remaining images were categorized as invalid. %(either all users were incorrect or human users could only correctly classify with low confidence suggesting they were correct by chance).
    Invalid images revealed unanticipated behaviour of adversarial attacks, adding to images instances of objects with incorrect labels (``Deceptive'') and making the labeled instance imperceptible (``Ambiguous'').
    The ground truth classes are below\footnotemark.
    %Inspection of adversarial images incorrectly classified by both human users and GPT4 revealed unanticipated behaviour of adversarial attacks, notably, interpolation between classes, imperceptible resolution, or invalid images seemingly containing no object from the representative class.
    %This study and its findings support the irreducible error on the current CIFAR10 adversarial benchmark, as suggested by our scaling law.
    Clean images are provided in Appendix~\ref{sec:invalid_app}. 
    }
    \label{fig:human-gpt-comparison}
\end{figure}

\iffalse
\begin{figure}[t]
    \centering
    \includegraphics[width=\columnwidth]{figs_and_tables/FlowChart.pdf} 
    \caption{\textbf{\textcolor{red}{Flow Chart of categorizing data between the valid and invalid group.}}}
\end{figure}
\fi

\subsection{Addressing Invalid Data}
%and further suggests that roughly $10$\% of the test images are invalid when measuring the adversarial robustness of a model.
The inability of human users to correctly classify $\sim$$10$\% of adversarial images suggests that the $l_{\infty}$ attacks performed on our SOTA model produce invalid images, lacking a perceptible depiction of the labeled class. 
Naively, we could define invalid images as the set of images that are never correctly guessed, but this would not reflect that some guesses are correct by chance. 
Indeed, we find that humans who lack confidence in recognizing an instance of the labeled class in the image are often correct nevertheless (see Figure \ref{fig:adv-clean-by-human-conf}). 
Thus, to reduce the extent to which we label as ``valid’' the images that are correctly guessed by chance, we define valid images as the $1902/2629$ images correctly labeled by $2/4$ users or $1$ \textit{human} user %who has high confidence about their guess
with high confidence---users with high confidence are typically correct and likely see an instance of the label class in the image (Figure~\ref{fig:adv-clean-by-human-conf}). The other $727/2629$ images are categorized as invalid.

%To better understand adversarial misalignment and the resulting irreducible error, we analyze data from our study and partition images into two categories 
In Figure~\ref{fig:human-gpt-comparison}, we show examples of \textbf{valid} and \textbf{invalid} images. 
We also illustrate two flavors of invalid images: (i) \textit{deceptive} images that depict an object with an incorrect class, deceiving users; and (ii) \textit{ambiguous} images that depict no object clearly, confusing users. 
\textit{Deceptive} images, representing $39$\% of invalid images, are those causing our SOTA NN and a majority of humans to make the same incorrect prediction.\footnotetext{1:cat, 2:bird, 3:deer, 4:horse, 5:deer, 6:cat, 7:dog, 8:bird, 9:bird, 10:cat, 11:airplane, 12:deer, 13:cat, 14:horse, 15:airplane}
%Valid images were those that were either ($i$) correctly classified by at least half of all users (including GPT-4 as a user) regardless of confidence or ($ii$) correctly classified with high confidence by only one human user. 
%GPT-4 predictions were not considered for ($ii$) due to the observed unreliability of GPT-4 confidence. 
%All remaining images were categorized as invalid. %since either ($i$) all human users were incorrect, or ($ii$) human users could only correctly classify these images with low confidence (suggesting that they were correct by chance).
%Images in the invalid category revealed unanticipated behaviour of adversarial attacks such as interpolation between classes and imperceptible images due to low resolution.

Interestingly, prior work shows that visually ambiguous or illusory images can be interpreted in multiple ways by both humans and machines, e.g. depending on the prompt given to a generative classifier \citep{jaini2023intriguing}. Uniquely, we show that data generated by adversarial attacks harms performance not just by creating ambiguity that classifiers fail to deal with, but rather it can also create a clear depiction of an incorrect class that deceives both human and machine users into confidently making the same incorrect prediction. Indeed, visual inspection of the \textit{deceptive} images, which make up 39\% of all invalid images, illustrates clear depictions of incorrect classes (e.g., see the images in the top row of Figure \ref{fig:human-gpt-comparison} and the right side of Figure \ref{fig:teaser}).

\begin{table}
\caption{Benchmarking has a new performance limit when it excludes invalid data. Our SOTA NN's performance with and without invalid data is shown in the first row. We also estimate upper bounds on average human performance: the first column of the second row adds the $7371$ images correctly classified by the SOTA NN to the average number of images correctly classified by humans when they are given the $2629$ images misclassified by our SOTA NN, then divides by $10000$. Removing invalid data moves the estimated average human performance closer to $100$\%.}
\label{tab:update-bench}
\vskip 0.1in
\centering
%\tiny
%\resizebox{\columnwidth}{!}{%
%\scalebox{0.7}{
\begin{widetabular}{\linewidth}{lcc}
\toprule
%& \multicolumn{2}{c}{\textbf{Benchmarking Approach}} \\ 
%Configuration & Original & Updated \\ 
Configuration & All Data & Valid Data \\ 
\midrule
SOTA NN & ${73.71}$\% & {${79.49}$\%} \\ %(7371)/(10000-727)
%SOTA NN with GPT-4 & $85.31$\% & $91.04$\% \\
SOTA NN $+$ Avg Human & $90.46$\% & $96.46$\% \\
%SOTA NN with Best Human & $92.77$\% & $98.18$\% \\
\bottomrule
\end{widetabular}
%}
\end{table}

To better track robustness as a fraction of human performance, we remove from the CIFAR10 $\ell_{\infty}$ adversarial benchmark test set \citep{croce2020robustbench} the images that AutoAttack can make invalid.
Table~\ref{tab:update-bench} shows our SOTA NN's robustness on the new and original benchmarks. Clarifying the limitations imposed by invalid images, Table~\ref{tab:update-bench} also shows the SOTA NN's $7371$ correct classifications plus the average user's correct classifications on the $1902$ ($2629$) valid (original) images remaining, divided by $9273$ ($10^4$). 

\section{Discussion}
\label{sec:discussion}
By deriving scaling laws for CIFAR10 adversarial robustness, we showed that compute-efficient configurations (i.e., model and dataset sizes), better synthetic data generators, and more compute can all help improve adversarial robustness. 
Indeed, we improve on the prior SOTA by $3$\%, reaching $73.71$\% AutoAttack accuracy. 
However, we also find %in our scaling laws 
that there are asymptotic limits to performance: it appears that some adversarial data will never be correctly classified. To address this surprising finding, we evaluate human performance on the adversarial data that our SOTA NN misclassifies, and we find that humans fail to classify much of this data too. In other words, when applied to NNs with SOTA robustness, existing attack formulations that bound attacks using $\ell_p$-norm constraints generate adversarial data that is \textit{invalid}. This finding is consistent with prior works that criticize the ability of the $\ell_p$ norm to quantify perceptual similarity~\citep{sharif2018suitability, tversky1977features}. Invalid data prevents benchmarking from returning a fraction of human-level robustness. Accordingly, we argue for designing an attack formulation that takes image validity into account.

%removing invalid data from benchmarking, and we produce a revised CIFAR10 benchmark dataset without images that are known to become invalid after adversarial attacks.

%We investigate invalid adversarial data and find that it comes in two flavors: data depicting an instance of the incorrect class, and data depicting no perceptible instances of any class. Since invalid data cannot be correctly classified by humans, it prevents benchmarking from returning the fraction of progress towards human-level robustness. Accordingly, we argue that invalid data should not be used to benchmark neural network robustness, and we produce a revised CIFAR10 benchmark dataset that does not contain images that are known to become invalid after adversarial attacks.

\subsection{Limitations and Future Directions}

We sought to clarify adversarial robustness limits suggested by the \textbf{unsolved state} of a well-studied, small-scale adversarial robustness problem: $\ell_{\infty}$-norm-constrained AutoAttack on CIFAR10 \citep{croce2020robustbench}. 
In doing so, despite the large scale of our experiments (over $10^{22}$ total FLOPs and hundreds of training runs), we made observations that need to be tested more broadly. 
For instance, prior work suggests that invalid data arises on higher resolution datasets like ImageNet \citep{gaziv2023robustified}, but the extent to which invalid data affects performance on other robustness benchmarks requires further study. 
Future work can extend ours by exploring other impacts of invalid data on the robustness problem. 
We focused primarily on the effect of invalid data that arises during benchmarking, but our preliminary results show that invalid data can be created by training attacks too (Appendix \ref{sec:misalign-training}).
We also made observations about robustness scaling behavior that should be tested with other datasets and models: studying WideResNets, we found that $N$ and $D$ should be scaled at roughly the same rate when synthetic data quality is high -- it will be interesting to examine this more generally (e.g. with ViTs on ImageNet).

Our human study was limited to three users. To address this limitation, we make a subset of this study's adversarial data publicly available via an online quiz, and we will make more data available upon request, allowing the community to contribute to the estimates we made in Section \ref{sec:human}.
Finally, while we provide an updated, more optimistic view of progress towards human-level robustness on CIFAR10, our scaling laws also suggest that the costs of reaching human performance are impractically high. 
More efficient algorithms \citep{bartoldson2023compute} for adversarial training can address these costs and open up paths to robust models.
%github.com/bbartoldson/Adversarial-Robustness-Limits

\section*{Impact Statement}
This paper presents work whose goal is to advance the field of Machine Learning. There are many potential societal consequences of our work, none which we feel must be specifically highlighted here. Our organization’s IRB deemed our work to have no need for an IRB review.


\section*{Acknowledgements}
We thank Nicholas Carlini, Maksym Andriushchenko, Vikash Sehwag, and Pin-Yu Chen for helpful comments. We thank Youngsoo Choi for facilitating our human evaluation. This work was performed under the auspices of the U.S. Department of Energy by Lawrence Livermore National Laboratory
under Contract DE-AC52-07NA27344 and LDRD Program Project No. 23-ERD-030 (LLNL-CONF-860190).
%}

\bibliography{references}
\bibliographystyle{icml2024}

%%%%%%%%%%%%%%%%%%%%%%%%%%%%%%%%%%%%%%%%%%%%%%%%%%%%%%%%%%%%%%%%%%%%%%%%%%%%%%%
%%%%%%%%%%%%%%%%%%%%%%%%%%%%%%%%%%%%%%%%%%%%%%%%%%%%%%%%%%%%%%%%%%%%%%%%%%%%%%%
% APPENDIX
%%%%%%%%%%%%%%%%%%%%%%%%%%%%%%%%%%%%%%%%%%%%%%%%%%%%%%%%%%%%%%%%%%%%%%%%%%%%%%%
%%%%%%%%%%%%%%%%%%%%%%%%%%%%%%%%%%%%%%%%%%%%%%%%%%%%%%%%%%%%%%%%%%%%%%%%%%%%%%%
\newpage
\appendix
\onecolumn
\section{Outline of Our Contributions and Their Implications}
\label{sec:contributions}
Here we present a more detailed list of our contributions and their implications. 

Our contributions are as follows:
\begin{itemize}
    \item We derive the first scaling laws for CIFAR10 adversarial robustness that take synthetic data quality into account (Sections \ref{sec:approach_2} and \ref{sec:approach_3}).

    \item Using data generated from large-scale experimentation (238 training runs and over $10^{22}$ FLOPs), we show that our scaling laws can accurately model and predict robustness performances across various unseen model sizes, dataset sizes, and synthetic data qualities (Section \ref{sec:scaling_laws}).

    \item We compute: (a) the optimal resource allocation (i.e., model and dataset size) at various training FLOPs values, allowing robustness to be maximized given a budget; and (b) the optimal return on investment, i.e., the best robust loss/accuracy that can be achieved for a given compute budget (Equation \ref{eq:n_star_d_star} and Figure \ref{fig:all_n_stars} for (a), and Figure \ref{fig:teaser} for (b)).

    \item We also derive the tradeoff between model size and training compute overhead, showing that there are vast opportunities to reduce model size (improve inference speed) while actually decreasing training compute overhead (Equation \ref{eq:overhead} and Figure \ref{fig:overhead}).

    \item We carry out a small-scale human study to gain a deeper understanding of the performance bottleneck of current adversarial training methodologies (Section \ref{sec:human}).

    \item Additionally, we are the first to perform a single-epoch (i.e., seeing each image only once), synthetic-data-only adversarial training run that outperforms the prior state-of-the-art, which relied on using non-synthetic data (Figure \ref{fig:all_n_stars}).

\end{itemize}

Via our scaling law and human studies, we provided actionable insights to advance the field of adversarial robustness:

\begin{itemize}
    \item With the compute-optimal settings from our scaling laws, we showed that the prior SOTA adversarial training run \cite{wang2023better} used a model too large for its training FLOPs budget, using more compute than necessary to achieve its robustness level, and we matched this prior robustness peak with 20\% fewer training FLOPs (Figures \ref{fig:all_n_stars} and \ref{fig:overhead}).

    \item We trained several new models in a compute-efficient manner to make a significant leap in SOTA robustness -- our best model beats the prior SOTA by 3\% AutoAttack accuracy (Figures \ref{fig:teaser} and \ref{fig:all_n_stars}). 
    
    \item A model predicted to be exactly compute-optimal at the prior SOTA’s FLOPs budget \cite{wang2023better} outperformed the prior SOTA by $1$\% AutoAttack while being $3\times$ smaller (FLOPs and parameters), boosting training performance and inference efficiency (Figure \ref{fig:overhead}).

    \item We found that ``scaling up compute" alone is not an effective strategy to achieve a human robustness level, as our scaling laws predict that current approaches will need  thousands of years of state-of-the-art supercomputer time to reach this level (Figure \ref{fig:teaser}).
    
    \item We also found that human performance itself on this task is far from perfect: existing attack formulations (relying on an $\ell_{\infty}$-norm bound to ensure attacks preserve the original label) concerningly allow generation of invalid images that do not fit their original label (Figures \ref{fig:teaser}, \ref{fig:adv-clean-by-human-conf}, and \ref{fig:human-gpt-comparison}).

\end{itemize}

Some broader implications of our work for the adversarial robustness area are as follows:

\begin{itemize}
    \item Advances on the adversarial robustness problem will require the design of more efficient training algorithms and improved architectures, rather than simply scaling (Section \ref{sec:discussion}).
    
    \item Attack formulations must be rethought to solve this problem; e.g., attacks should only produce valid images that contain the information needed to assign the original label (Section \ref{sec:human}).

    \item Robustifying models closer to human levels makes them greater threats for misuse, as they can be combined even with naive attack approaches ($\ell_{\infty}$-norm-bounded attacks) to create adversarial examples that transfer to humans, fooling them and creating potential security concerns (Section \ref{sec:human}).

\end{itemize}

\section{Counting the FLOPs of Adversarial Training}
\label{sec:FLOPs}

We use the PyTorch \citep{paszke2019pytorch} flop counter to find the FLOPs consumed by adversarial training. We find that our adversarial training loop requires $27\times$ the FLOPs required to complete $1$ forward pass, whereas regular training is typically estimated to require about $3\times$ the forward pass FLOPs. The added cost of adversarial training is primarily due to the 10 PGD iterations used to adversarially perturb images at training time. The following explains this in detail.

\subsection{Why is adversarial training $9\times$ as expensive per iteration?}

We clarify why an adversarial training iteration, using 10 PGD steps and the TRADES loss, requires $\sim$$9\times$ the FLOPs of a regular training iteration. First, we note that a typical training iteration uses $1$ forward pass and $1$ backward pass, which is $\sim$$2\times$ as FLOP-intensive as a forward pass. We approximate this as $3$ forward passes of compute. Next, we show that a typical adversarial training iteration requires $27$ forward passes of compute. 

Creating an adversarial image for the TRADES loss requires a PGD attack that only computes the gradient with respect to the image (not the weights), making the backward pass for each step of this attack only $\sim$$1\times$ as FLOP-intensive as a forward pass. Thus, for 10 PGD steps, we use $\sim$$20$ forward passes of compute. This number rises to $21$ because a forward pass on the clean data is used to obtain the clean-image logits used in the PGD attack. 

Two additional forward passes are used to compute the TRADES loss, which compares the output of the model on the attacked and clean data (a more efficient implementation could save a forward pass by reusing the clean data output needed by the PGD attack). Finally, there is a backward pass used to compute the gradient of the TRADES loss. This backward pass is $\sim$$2\times$ as large as a typical backward pass because the batch size is effectively doubled via the loss function's use of model outputs from two different inputs. Thus, given a PGD-attacked image, the TRADES loss requires $6$ forward passes of compute, making the entire adversarial training iteration require about $\sim$$27$ forward passes of compute.

\section{Additional Discussion of Adversarial Data That Fools Humans}
\label{sec:tricking_humans}
\citet{elsayed2018adversarial} generate adversarial examples by attacking CNNs and show that human accuracy falls on these perturbed images when the humans are only briefly exposed to the images and asked to quickly classify the image. While \citet{elsayed2018adversarial} find that human predictions are generally not harmed by their study's small adversarial perturbations when the humans are given sufficient time to make their choice, we find that adversarial data created by common attack approaches do fool humans, making that adversarial data's human label misaligned with the label used to compute accuracy. 

Our finding that perturbations that cause artificial network misclassification often also cause human misclassification may not be surprising. Adversarial robustness comparisons of humans and artificial neural networks are further explored by \citet{subramanian2023spatial}, who show that humans make classifications using a narrower range of frequencies within the range of frequencies relied upon by artificial networks for classification decisions. Thus, perturbations that fool neural networks may act on frequencies that humans rely on for classification, potentially explaining the results of our human-NN comparison study. This is consistent with human comments after the study: e.g., ``the attack was able to manipulate the image in some way that it made it very confusing (either changing semantics, blurring part of the object/class into the background, etc.)''.

\citet{guo2022adversarially} show the susceptibility of biological, primate neurons to adversarial attacks, indicating that use of adversarial training that aligns artificial neural network performance with human performance may not produce models impervious to adversarial attack. Our work is consistent with the argument that making artificial models more human-like will not solve the adversarial robustness problem, as we observe that images that fool state-of-the-art neural networks also fool humans. Indeed, our results show that measures of progress on the adversarial robustness problem may be too pessimistic -- i.e., by showing that humans often fail to classify adversarially perturbed CIFAR10 images, we clarify that neural networks are closer to human-level (``gold-standard'') performance than previously thought.

\newpage
\section{Hyperparameter Study}
\label{sec:hparam}

\begin{figure}[h]
    \centering
    \includegraphics[width=0.5\linewidth]{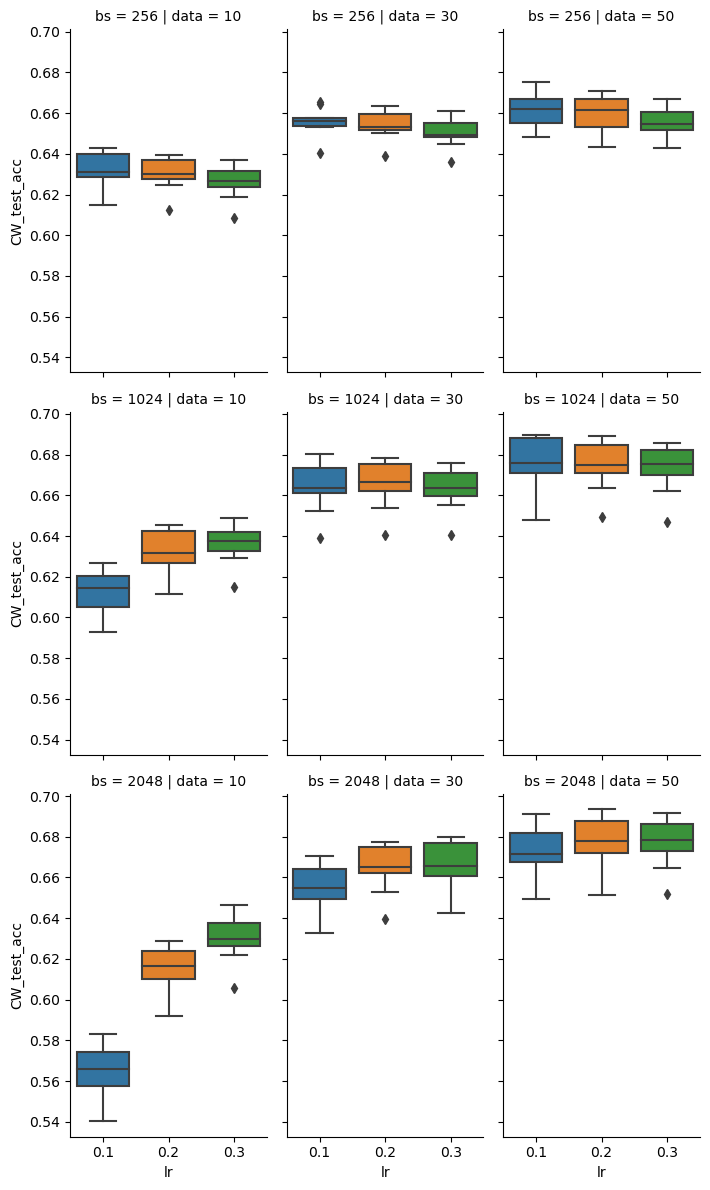}
    \caption{\textbf{Hyperparameter study.} We trained a variety of preliminary models on different dataset sizes (``data'' in the plot, shown in millions) in order to understand how to set key hyperparameters like batch size (``bs'' in the plot) and learning rate (``lr'' in the plot) prior to training models for our scaling laws. Settings with higher test accuracy after a PGD attack on the CW loss (``CW\_test\_acc'' in the plot) were chosen for our scaling law training runs.}
    \label{fig:hparams}
\end{figure}

Prior to training models used to fit our scaling laws, we conducted a hyperparameter study to understand how to set key hyperparameters as a function of model size and dataset size. A sample of our findings from this study is shown in Figure \ref{fig:hparams}. We use such results to support the settings we discuss using in our main text, which we reiterate here: datasets with 10M or fewer samples use batch size 1024, otherwise batch size is 2048; the learning rate is 0.3 for datasets with 10M or fewer samples, 0.2 for larger datasets with up to 200M samples, and 0.1 for datasets with 300M samples.

\newpage
\section{Training Dataset Sampling}
\label{sec:data}

\begin{figure}
    \centering
    \includegraphics[width=1\linewidth]{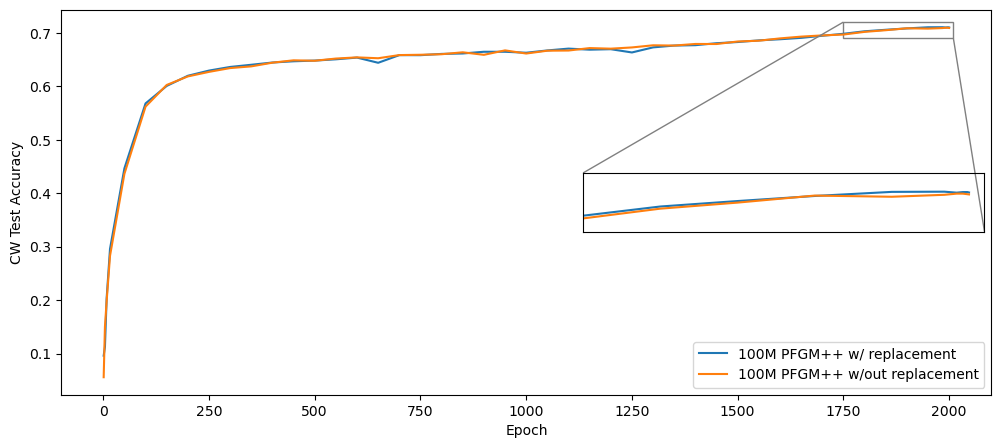}
    \caption{\textbf{Sampling with and without replacement are similar.} Infrequent resampling of the same sample that sometimes occurs when training models used to fit our scaling laws has no notable effect on performance.}
    \label{fig:resample}
\end{figure}

When training on synthetic data, we aim to see each sample a single time. However, we sometimes see a specific sample more than once during a training run for three reasons:
\begin{enumerate}
    \item Our data sampler uses a constant batch size for each training step, causing the actual number of samples seen per epoch to be slightly above the \num{50000} typical for CIFAR10 training data. Our training epochs calculation is made assuming \num{50000} samples are seen each epoch, so we start to reuse synthetic samples seen at the start of training to finish the final few epochs. For example, assume we see \num{51200} samples per epoch: if training on $5$M samples, we would train for $100$ epochs and revisit $100 \times (\num{51200} - \num{50000}) = \num{120000}$ samples. All other samples would be seen once during training.
    
    \item Training the SOTA robustness model requires 500 million high-quality synthetic examples, and we train on EDM-20 for 1.5 ``epochs'', DG for 1.5 ``epochs'', and PFGM++ for 2 ``epochs'' (reaching a total of 500 million examples). In this paragraph, ``epochs'' refers to passes through the full \num{100000000} element datasets. In most of the training runs discussed in Section \ref{sec:efficient}, we also mix in non-synthetic CIFAR10 data, so the actual number of full passes through each of these synthetic datasets is slightly lower.
    
    \item When training models used for our scaling laws, we sampled images without replacement (in a random order) but did not reload the data sampler's state when interrupted jobs resumed. This leads to some amount of sampling \textit{with} replacement, which was the strategy used by \citet{wang2023better}. In particular, when training was interrupted by a job time limit (every 12 hours), our training would resume with a new (random) data sampler state, causing some previously seen examples to be resampled. We show that this form of infrequent resampling causes no notable difference relative to training without replacement (by reloading the data sampler state) in Figure \ref{fig:resample}. The models we train in Section \ref{sec:efficient} reload the data sampler state to ensure sampling without replacement.
\end{enumerate} 

Similar to \citet{wang2023better}, we see non-synthetic CIFAR10 examples multiple times during training when mixing together original CIFAR10 and synthetic data in our training runs not used for scaling laws (Section \ref{sec:efficient}).

\section{Adversarial Scaling Laws: Additional Results and Details}
\label{sec:fitting}

%%%%%%%%%%%%%%%%%%%%%%%%%%%%%%%%%%%%%%%%%%%%%%%%%%%%%%%%%%%%%%%%%%%%%%%%%%%%%%%
%%%%%%%%%%%%%%%%%%%%%%%%%%%%%%%%%%%%%%%%%%%%%%%%%%%%%%%%%%%%%%%%%%%%%%%%%%%%%%%

\subsection{Approach 1}
\label{sec:approach_1_app} 

Given empirically observed $N^*, D^*$, the points on the envelope in Figure \ref{fig:approach1}, Approach 1 uses regressions in log space to find the optimal $N^*, D^*$ as power law functions of FLOPs. The results of these regressions are shown in Figure \ref{fig:approach1}. 

The query FLOPs values used to collect $N^*, D^*$ are obtained by taking a set $\mathcal{F}$ of 1000 logarithmically spaced values of training FLOPs, then restricting to the subset $\mathcal{Q} = \{ f_i \in \mathcal{F} \mid \nexists f_j \in \mathcal{F} \text{ such that } (Loss_j < Loss_i) \land (f_j < f_i) \}$. The restricted query points are those that reflect loss improvements when the training FLOPs budget increases; i.e., the restriction removes upticks in the envelope at regions with an insufficiently dense sampling of training dataset sizes, areas of Figure \ref{fig:approach1} where the envelope is flat. The optimal loss at each query training FLOP value produces the envelope seen in the black line of Figure \ref{fig:approach1}.

\begin{figure}
    \centering
    \includegraphics[width=1\linewidth]{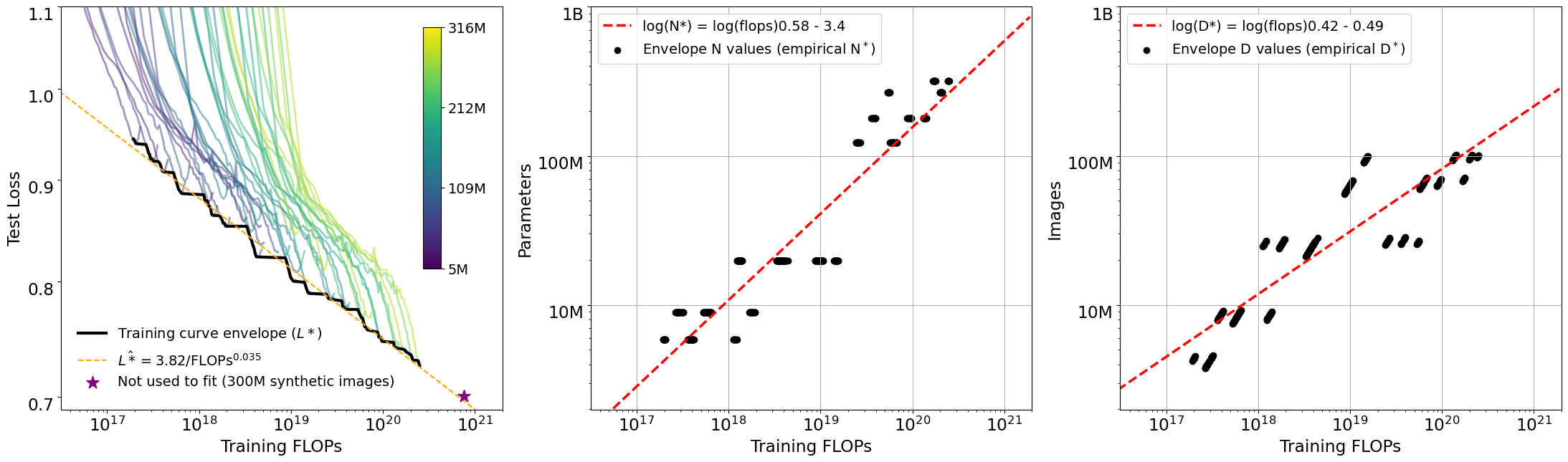}
    \caption{\textbf{Approach 1: Envelope of Training Curves.} Our findings in Figure \ref{fig:approach1} are stable when the training dataset is changed to DG.}
    \label{fig:approach1_dg}
\end{figure}

When using other datasets with high quality, we find similar results to those shown in Figure \ref{fig:approach1}. For instance, Figure \ref{fig:approach1_dg} shows similar results are obtained when using DG instead of PFGM++.

\subsection{Approach 2}
\label{sec:approach_2_app}

\subsubsection{Fitting}

We use L-BFGS to learn the parameters of Equation \ref{eq:approach_2}, reproduced below for convenience:
\begin{equation*}
\begin{aligned}
    &L(N,D) = \frac{A}{N^{\alpha}} + \frac{B'}{D^{\beta}} + E', 
    \\
    &E' = \mathrm{exp}(\mathrm{log}(E) + \mathrm{log}(1+\mathrm{Quality}^{-1})\epsilon), 
    \\
    &B' = \mathrm{exp}(\mathrm{log}(B) + \mathrm{log}(1+\mathrm{Quality}^{-1})\zeta).
\end{aligned}
\end{equation*}

Our approach largely follows that of \citet{hoffmann2022training}, with the exception that we also learn the values of $\epsilon, \zeta$. L-BFGS optimizes the Huber loss with $\delta=0.001$. The observed error we feed into the Huber loss is computed in the following way. We begin with candidate values for $a, b, e, \alpha, \beta, \epsilon, \zeta$. We define $a=\mathrm{log}(A)$. We transform the logs of $B, E$ ($b, e$ at FID=0) according to the data quality parameter; for example: $b = \mathrm{log}(B) + \mathrm{log}(1+\mathrm{Quality}^{-1})\zeta$. We then compute the log of the predicted loss: $\mathrm{log}(\hat{L}) = \mathrm{log\_sum\_exp}([a - \alpha\mathrm{log}(N), b - \beta\mathrm{log}(D), e])$. Finally, the error used in the Huber loss is  $\mathrm{log}(\hat{L})-\mathrm{log}(L)$.

When learning $A, B, E, \alpha, \beta$, we initialize L-BFGS with each element in the Cartesian product of the following initial points, ultimately selecting the values associated with the minimum achieved Huber loss. $a: [0,1,2,5,10]$, 
    $b: [0,1,2,5,10]$, 
    $e: [-1,-.5,0,.5,1]$
    $\alpha: [0,0.1,0.25,0.5,1]$,
    $\beta: [0,0.1,0.25,0.5,1]$.

When learning $\epsilon, \zeta$, we initialize L-BFGS with each element in the Cartesian product of the following initial points, ultimately selecting the values associated with the minimum achieved Huber loss. $\zeta: [-.3, -.15, .15, .3]$, $\epsilon: [0.01, 0.1, 0.2]$.

We obtain the following values:  

$A=6.69$, $B=9.89$, $E=0.48$, $\alpha=0.24$, $\beta=0.23$, $\epsilon=0.16$, $\zeta=-0.28$.

Note that we fit Equation \ref{eq:approach_2} to all of the data specified in Section \ref{sec:approach_2} with one exception: we do not fit to ``small'' models trained on ``large'' datasets. The specific restriction is that we drop training runs where the following conditions are all true: model size is less than 100M, data size is greater than 10M, FID is less than 10. We discuss the reason for this restriction in Section \ref{sec:approach_3_app}, and below we show that removing this restriction does not cause our Approach 2 fitted values to change much:

$A=7.6$, $B=9.70$, $E=0.49$, $\alpha=0.25$, $\beta=0.23$, $\epsilon=0.14$, $\zeta=-0.25$.

\subsubsection{Loss-accuracy regression}
\label{sec:loss_acc}

\begin{figure}
    \centering
    \includegraphics[width=0.5\linewidth]{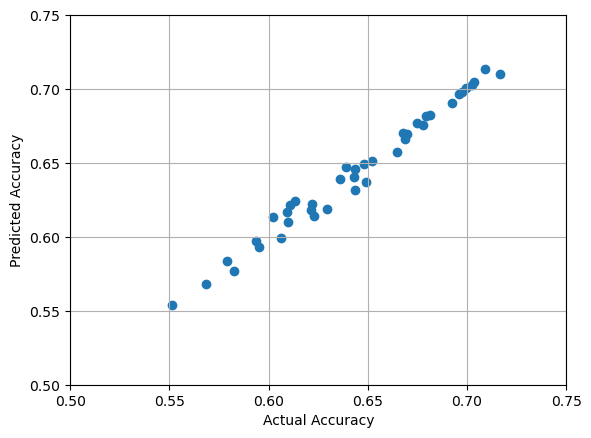}
    \caption{\textbf{Predicting accuracy from loss.} We create an accuracy predictor to facilitate conversion of predicted loss values to predicted accuracy values.}
    \label{fig:reg}
\end{figure}

In order to create Figure \ref{fig:teaser}, which predicts adversarial accuracy values, we convert our scaling law's loss predictions to accuracy predictions. To do this, we regress accuracy on loss using observations of models trained on the DG dataset, restricting to the subset of observations at which models obtain their minimum loss. Using this regression-based-predictor results in good agreement between predicted and actual values ($R^2=0.98$), as shown in Figure \ref{fig:reg}. The learned slope is $-0.7496$, the learned intercept is $1.2575$.

\subsection{Approach 3}
\label{sec:approach_3_app}

\subsubsection{Fitting}

We use L-BFGS to learn the parameters of Equation \ref{eq:approach_3}, reproduced below for convenience:
\begin{equation*}
\begin{aligned}
 &L(N,D,\mathrm{Quality}) = \frac{A}{N^{\alpha}} + \left(\frac{B}{D} + \left(\frac{Q}{\mathrm{Quality}}\right)^{(\kappa/\beta)}\right)^{\beta} + E',
 \\
    &E' = E + \mathrm{log}(1+\mathrm{Quality}^{-1})\epsilon .
\end{aligned}
\end{equation*}

\begin{figure*}[t]
    \centering
    \includegraphics[width=1\linewidth]{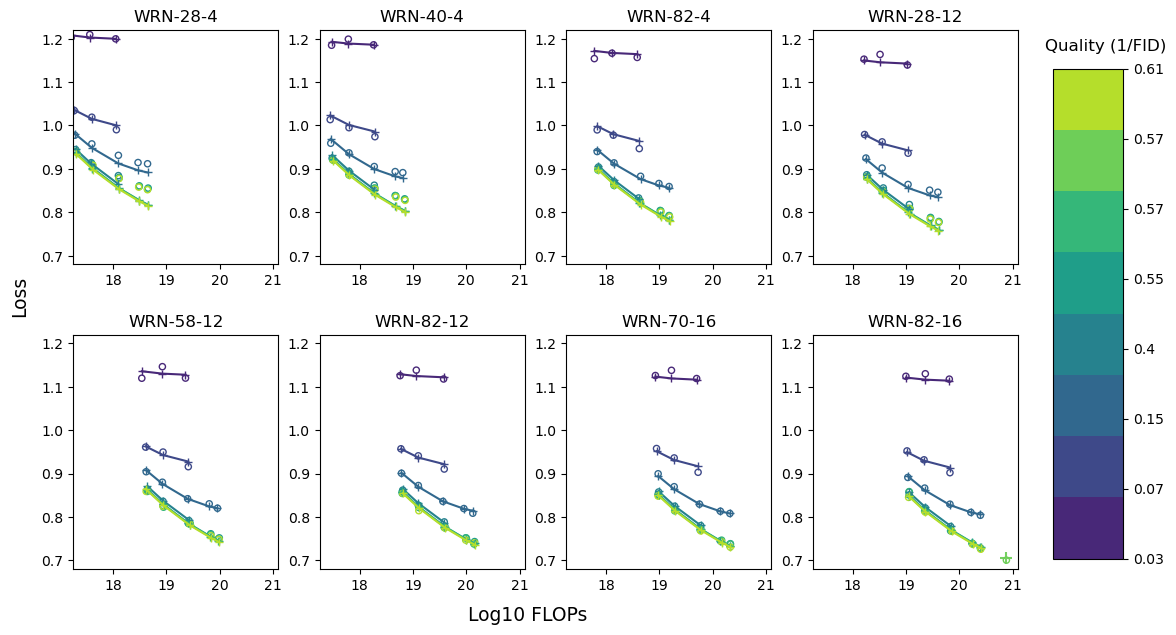}
    \caption{\textbf{Fit of Approach 3 on all models.} This is a version of Figure \ref{fig:approach3_fit_4_models} that shows the fit of Approach 3 to the other 4 models used to learn our scaling laws. Note that the same scaling law parameters are used in all 8 plots, demonstrating Approach 3's ability to simultaneously capture the effects of model, dataset, and quality scaling.}
    \label{fig:approach3_fit}
\end{figure*}

The fitting procedure largely mirrors the one used for Approach 2 described in Section \ref{sec:approach_2_app}, and we describe notable differences in this paragraph. When fitting Approach 3, in addition to using results from all of the datasets listed in Table \ref{tab:datasets}, we also use the performance of one model trained on the $300$M synthetic images from our three best generators (EDM-20, PFGM++, DG), which is visible in the bottom right plot of Figure \ref{fig:approach3_fit}. When learning $A, B, E, Q, \alpha, \beta, \kappa, \epsilon$, we initialize L-BFGS with each element in the Cartesian product of the following initial points, ultimately selecting the values associated with the minimum achieved Huber loss.
$A: [5,6,7]$, $B: [6500,7000,7500]$, $E: [.6, .5]$, $Q: = [0.01, 0.5]$, $\alpha: [0.1, .2, .3]$, $\beta: [0.1, .2, .3]$, $\kappa: [.8,.6]$, $\epsilon: [.01]$.

The specific values we learn are the following:

$A=6.0$, $B=7000.0$, $E=0.52$, $Q=0.007$, $\alpha=0.24$, $\beta=0.22$, $\kappa=0.6$, $\epsilon=0.04$.

In Figure \ref{fig:approach3_fit}, we show that these values produce good fits for all models, datasets, and qualities. Note that at FID=0, Equations \ref{eq:approach_2} and \ref{eq:approach_3} have similar forms and their values can be compared. The main exception to this is that $B$ in Approach 2 should be compared to $B^{\beta}=6.85$ in Approach 3.

We fit Equation \ref{eq:approach_3} to all of the data specified in Section \ref{sec:approach_3} with one exception: we do not fit to ``small'' models trained on ``large'' datasets. The specific restriction is that we drop training runs where the following conditions are all true: model size is less than 100M, data size is greater than 10M, FID is less than 10. We impose this restriction because small models display underfitting on larger, higher quality datasets. For an example of this, see the top left plot of Figure \ref{fig:approach3_fit}, where WRN-28-4 losses are above the predicted values at larger FLOP values (i.e., larger dataset sizes). Modeling this underfitting behavior with our scaling law is possible, but we opted instead to learn a scaling law that applies to models that are not trained on far too much data given their size, as current adversarial training works typically do not enter this regime (using large model sizes and smaller dataset sizes).

\begin{figure}
    \centering
    \includegraphics[width=1\linewidth]{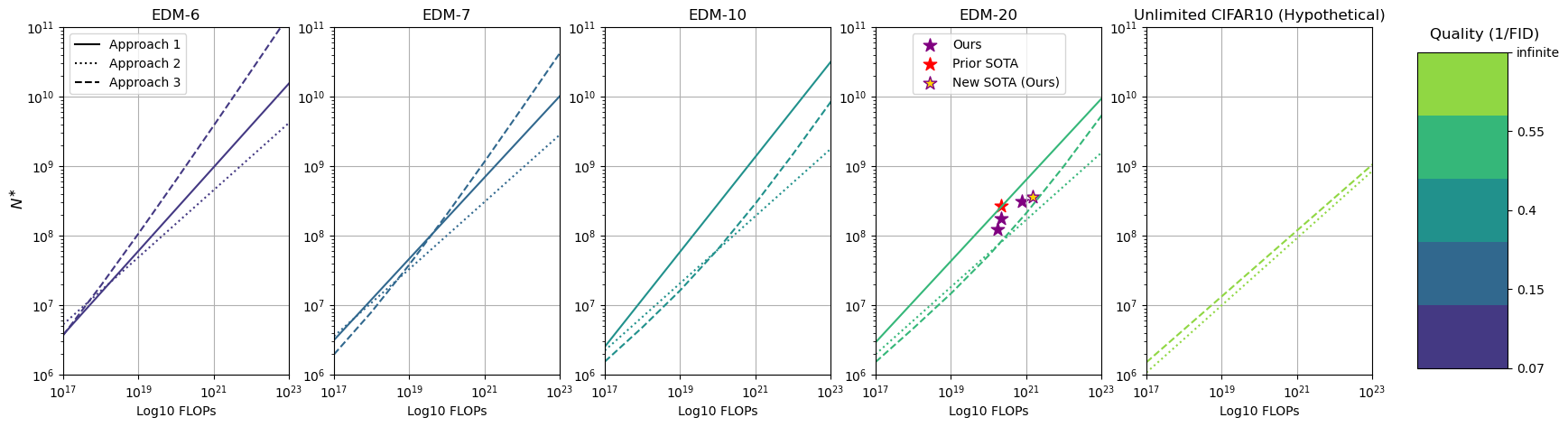}
    \caption{$\mathbf{N^*}$ \textbf{comparisons.} As data quality improves, our three scaling law approaches tend to predict smaller $N^*$. A version of the fourth plot is shown in the main text in Figure \ref{fig:all_n_stars}.}
    \label{fig:approach_compare}
\end{figure}

\subsubsection{Computing Optimal N, D}

Unlike Equation \ref{eq:approach_2}, the minimum of Equation \ref{eq:approach_3} does not have a simple analytical form. Accordingly, we find $N^*$ and $D^*$ using a numerical optimization approach (SciPy's `fsolve' routine).

\subsection{Compute-Efficient Adversarial Training}
\label{efficient_app}

\paragraph{Models and datasets used}

Here we provide additional model and dataset details for the results shown in Figures \ref{fig:all_n_stars} and \ref{fig:overhead}.

We train WideResNet-82-8 (79M parameters), WideResNet-58-12 (122M parameters), WideResNet-82-12 (178M parameters), WideResNet-82-16 (316M parameters), and WideResNet-94-16 (366M parameters). 

The 500M sample dataset used for our SOTA model (shown in Figure \ref{fig:all_n_stars}) is constructed from 300M unique samples as described in Appendix \ref{sec:data}.
The model trained on $150$M samples uses $50$M samples from each of our highest quality synthetic datasets (EDM-20, PFGM++, DG). For the model trained on $175$M samples, we use $25$M samples from this collection of $150$M samples twice. The model trained on $349$M samples used $100$M samples from each of our highest quality synthetic datasets, seeing $49$M samples twice. The model trained on $300$M samples used $100$M samples from each of our highest quality synthetic datasets and had no CIFAR10 data mixed in.

\paragraph{Comparing the Three Scaling Law Approaches}

In Table \ref{tab:approach_compare} and Figure \ref{fig:all_n_stars}, we see that Approach 1 favors scaling the model size more aggressively than Approach 2 ($a=0.56$ vs. $a=0.48$). Approach 3 represents a balance between the two approaches, shifting from suggesting the scaling of the model and dataset sizes at similar rates (at low training FLOPs values) to suggesting that additional compute should be primarily allocated toward increasing the model size due to its anticipating overfitting to the dataset of limited quality (at high training FLOPs values). Note that at the highest quality data (FID=0), Approach 3 suggests a constant $a,b$ similar to Approach 2's, as shown in the rightmost plot of Figure \ref{fig:approach_compare}.

All of our approaches suggest that the prior SOTA is trained on too little data given its model size. Thus, a central message of our results is to train smaller models on more data, consistent with the idea that robustness has a large sample complexity \citep{schmidt2018adversarially}.

\paragraph{Compute-Efficient Region}

In Figures \ref{fig:all_n_stars} and \ref{fig:overhead}, we use the DG-20 dataset to create the compute-efficient region, assuming that our training data will have FID $\approx1.65$. Due to the mixing in of non-synthetic CIFAR10 training data, the data quality available to the models shown in Figures \ref{fig:all_n_stars} and \ref{fig:overhead} could be slightly better than FID=1.65, which could mean the actual $N^*$ is even lower due to the tendency of higher quality data to raise $b$.

\section{Invalid Adversarial Data}
\label{sec:invalid_app}

\subsection{Human Evaluation}
\label{sec:human-eval}

\begin{figure}
    \centering
    \frame{\includegraphics[width=0.98\linewidth,trim={0 0 3mm 6mm},clip]{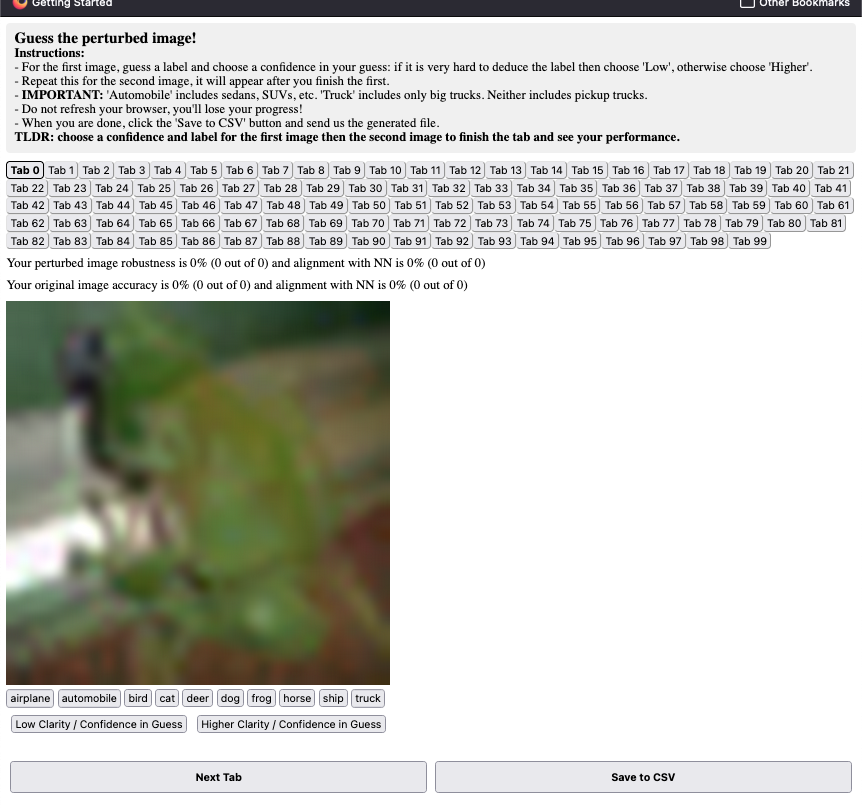}}
    \caption{The web application used to collect data in our human evaluations. Users were given instructions verbally, which were repeated at the top of the web page.}
    \label{fig:human_eval}
\end{figure}

To perform the human evaluation, we created a web application (see Figure \ref{fig:human_eval}). Users classified images in batches of 100, at their convenience in their own homes/offices over several days, and a unique web page was used for each batch of 100 images. Users classified the 2629 images that AutoAttack created which successfully tricked our SOTA NN, and they classified the clean versions of these images (thus, a total of 5258 images were classified by each of the 3 users).

The participants in our study were members of our research group. Unlike \citet{shankar2020evaluating}, we did not give users extensive training before they participated in this study. Being members of our research group, the users were familiar with CIFAR10. Further, CIFAR10 only has 10 classes, all of which are relatively distinct, reducing the importance of providing training/instruction beyond that which appears in the web application (shown in Figure \ref{fig:human_eval}).

\subsection{GPT-4 Evaluation}
\label{sec:gpt-prompt}
For completeness, we provide the prompt used for GPT-4 evaluation of adversarial images in Figure~\ref{fig:gpt-prompt}.
Additionally, we provide the payload used for GPT-4 in Figure~\ref{fig:gpt-payload}.
It is important to note that while GPT-4 is able to accurately correctly classify 1160 out of the 2629 adversarial images, this does not imply that GPT-4 is more adversarially robust than our SOTA model on the CIFAR10 $\ell_{\infty}$ benchmark, as the adversarial images evaluated by GPT-4 we generated via attacks on our SOTA model.
GPT-4 was selected as a fourth used for evaluation of the adversarial images due to its ability to understand subtle image details and its ability to understand the task via prompting \citet{yang2023dawn}.
However, this evaluation of GPT-4 on these adversarial images does suggest that adversarial attacks on the WideResNet-94-16 architecture, used for our SOTA model, are not fully transferable to GPT-4.

Trying numerous prompts would facilitate the optimization of GPT-4V performance and thus a potentially better understanding of which images truly have the information about the label destroyed by adversarial attacks. Prior to settling on the prompt we chose, we conducted a preliminary study that explored on a subset of images a prompt that expressed to GPT-4V that some images may have been attacked (and how that might affect object appearances), which led to a small improvement in performance ($\sim$5\%). Given the marginal improvement in performance and the significantly enhanced complexity of the prompt (that provided details the human users were not explicitly given), we opted for the prompt shown in Figure \ref{fig:gpt-prompt}.

%\begin{lstlisting}
%[language=Python, caption={GPT-4 Prompt for Classifying Adversarial Images}, label={lst:gpt4-prompt}]
\begin{figure}
\centering
%\begin{verbatim}
\begin{BVerbatim}
prompt = ("I will give you an image that contains one of 10 classes. "
          "The 10 classes are [airplane, automobile (but not truck 
           or pickup truck), bird, cat, deer, dog, frog, horse, 
           ship, and truck (but not pickup truck)]. "
          "Your reply to the image will tell me which of the 10 
           classes the image contains. "
          "Feel free to reason about the contents of the image, but the 
           final two words of your response must be in the following 
           format: '{class}, {confidence}'. "
          "'class' must be one of the 10 classes. 'confidence' must be 
           'low' or 'higher' -- only say 'low' if you think it is 
           unlikely that you have identified the correct class.")
\end{BVerbatim}
%\end{verbatim}
\caption{GPT-4 Prompt for Evaluating Adversarial Images.}
\label{fig:gpt-prompt}
\end{figure}
%\end{lstlisting}

\begin{figure}
\centering
\begin{BVerbatim}
payload = {
      "model": "gpt-4-vision-preview",
      "messages": [
        {
          "role": "user",
          "content": [
            {
              "type": "text",
              "text": prompt
            },
            {
              "type": "image_url",
              "image_url": {
                "url": img_base64
              }
            }
          ]
        }
      ],
      "max_tokens": 3200,
      "temperature": temperature
    }
\end{BVerbatim}
\caption{GPT-4 Payload for Evaluating Adversarial Images. ``temperature'' is set to 0, and ``img\_base64'' is the Base64 representation of the image.}
\label{fig:gpt-payload}
\end{figure}

\subsection{Clean versions of Figure~\ref{fig:human-gpt-comparison}}
\label{sec:clean-invalid-valid}
\begin{figure}[t]
    \centering
    \includegraphics[width=1.07\linewidth]{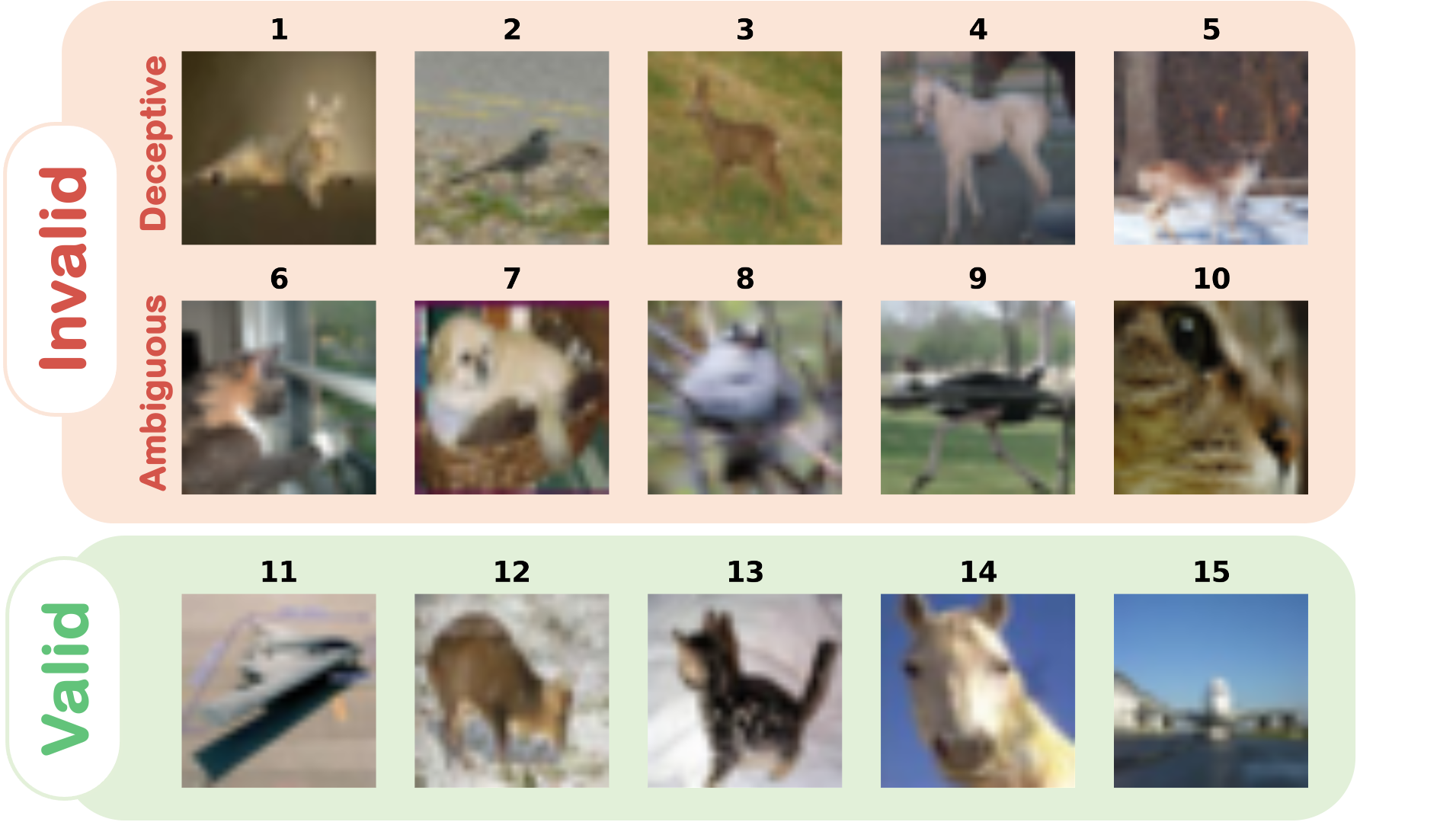} %,trim={10mm 72mm 10mm 0},clip]
    \caption{\textbf{Clean versions of adversarial images from Figure~\ref{fig:human-gpt-comparison}.} 
    %By saving the $2629$ test images that AutoAttack successfully attacked on our SOTA model, we were able to perform a human and GPT4 evaluation on the adversarial images. 
    We categorized 727 of the 2629 adversarial images as invalid ($27.87\%$ of images) based on analysis of our human and GPT-4 evaluations.
    Here we provide the original clean versions of the adversarial images from Figure~\ref{fig:human-gpt-comparison} with the same categorization and numbering.
    %Valid images were those correctly classified by at least half of the users (including GPT-4 as a user) or correctly classified with high confidence by exactly one human. %(as GPT-4 confidence was unreliable).
    %All remaining images were categorized as invalid. %(either all users were incorrect or human users could only correctly classify with low confidence suggesting they were correct by chance).
    %Invalid images revealed unanticipated behaviour of adversarial attacks such as interpolation between classes and images with imperceptible resolution.
    %Images were randomly sampled from each group for use in this figure and 
    The ground truth classes are: 1:cat, 2:bird, 3:deer, 4:horse, 5:deer, 6:cat, 7:dog, 8:bird, 9:bird, 10:cat, 11:airplane, 12:deer, 13:cat, 14:horse, 15:airplane. The NN predicted the following classes on the adversarial versions of these images: 1:bird, 2:deer, 3:bird, 4:dog, 5:dog, 6:deer, 7:frog, 8:frog, 9:airplane, 10:deer, 11:ship, 12:cat, 13:dog, 14:cat, 15:ship.
    }
    \label{fig:clean-human-gpt-comparison}
\end{figure}

In Figure~\ref{fig:clean-human-gpt-comparison}, we provide the clean (i.e., unattacked) versions of the images provided in Figure~\ref{fig:human-gpt-comparison} with the same invalid and valid categorization.
Of note, the classes of the images in the ``Deceptive" subcategory are much clearer and very different than the perceived classes by humans and our SOTA NN.
Specifically, perturbations in the "Deceptive" adversarial images in Figure~\ref{fig:human-gpt-comparison} appear to produce attributes of birds, deer, or dogs that are clearly not present in the original images.
Additionally, humans found the perturbations that were applied to the ``Ambiguous" adversarial version of the images made them appear warped or imperceptible while the clean images were easier to interpret and classify.

\subsection{Adversarial Training Images Can Be Invalid} 
\label{sec:misalign-training}
\begin{figure}
    \centering
    \includegraphics[width=1\linewidth]{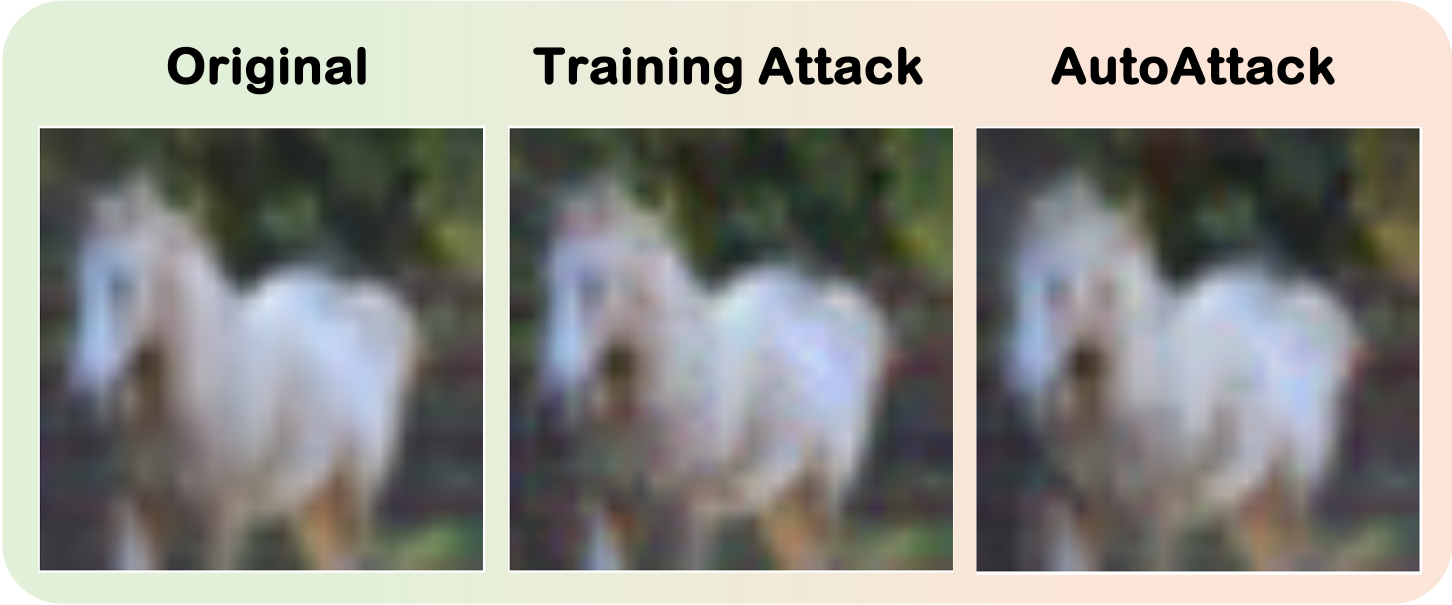}
    \caption{\textbf{Invalid training data.} While weaker than benchmark attacks, training attacks like 10-step PGD can produce adversarial data that fools humans. The middle image has the label ``horse'' but may appear as a ``dog''.}
    \label{fig:train_attack}
\end{figure}

%Removing invalid adversarial images from the benchmarking dataset can help benchmarking measure alignment of human and neural network performance on adversarial data. 
The ability of benchmark attacks to make 10\% of CIFAR10 adversarial test images invalid suggests that weaker, training-time attacks may generate invalid data too. In Figure \ref{fig:train_attack}, we show the original and AutoAttack-perturbed images from Figure \ref{fig:teaser} alongside the original image after the 10-step PGD attack used for training. Interestingly, the training attack results in an image (Figure \ref{fig:train_attack}, middle image) that appears to humans to be either a dog or a horse, or an interpolation between the two. The existence of invalid training images suggests that invalid adversarial data could also be removed at training time, rather than just during benchmarking.

\subsection{Adversarial Images That Can Fool Humans} 
\label{sec:fool-humans}
In Figure~\ref{fig:fool-humans}, we shown adversarial images that can potentially fool humans along with their clean counterparts.  

\begin{figure}
    \centering
    %\hspace*{-0.25\textwidth} % Adjust the value to shift the image to the left
    %\begin{minipage}{1.5\textwidth}
    %    \includegraphics[width=\linewidth]{figs_and_tables/clean-adversarial.png}
    %\end{minipage}
    %\vspace{-0.5 in}
    \includegraphics[width=\linewidth]{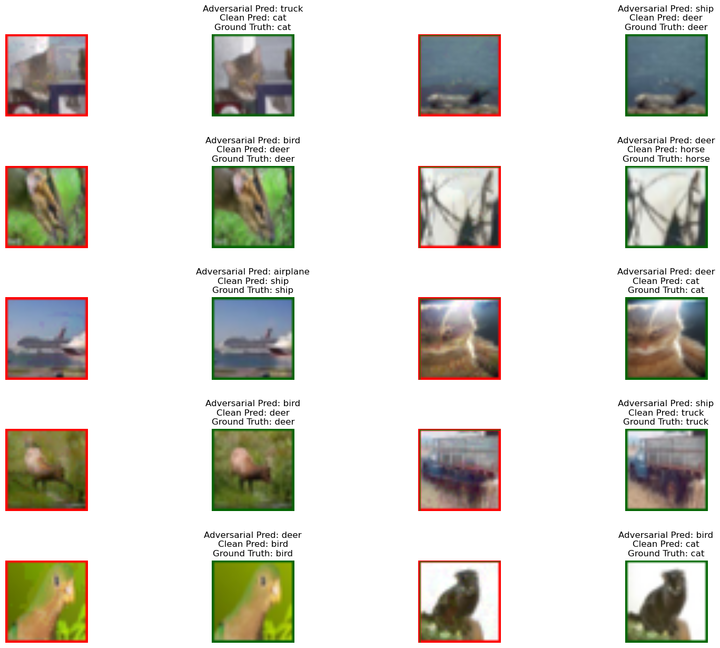}
    \caption{\textbf{Sophisticated manipulation strategies emerge with large-scale adversarial training, even when using naive attack algorithms.} Notice the subtle yet effective manipulations of clean images that can, in many cases, fool the human visual system. $\ell_\infty$ AutoAttack identifies the important parts of the image and then applies a combination of adversarial tactics, such as edge and texture tempering, discoloration, etc., to create effective adversarial examples. Adversarial examples from Figure \ref{fig:adversarial} are bounded by red boxes and reproduced here for convenience, their clean/original versions are bounded by green boxes.}
    \label{fig:fool-humans}
\end{figure}

%%%%%%%%
\iffalse
{
\begin{figure}
    \centering
    %\hspace*{-0.25\textwidth} % Adjust the value to shift the image to the left
    %\begin{minipage}{1.5\textwidth}
    %    \includegraphics[width=\linewidth]{figs_and_tables/clean-adversarial.png}
    %\end{minipage}
    %\vspace{-0.5 in}
    \includegraphics[width=\linewidth,trim={0 0 90mm 0},clip]{figs_and_tables/fig-21-update.png}
    \caption{\textbf{Sophisticated manipulation strategies emerge with large-scale adversarial training, even when using naive attack algorithms.} Notice the subtle yet effective manipulations of clean images that can, in many cases, fool the human visual system. $\ell_\infty$ AutoAttack identifies the important parts of the image and then applies a combination of adversarial tactics, such as edge and texture tempering, discoloration, etc., to create effective adversarial examples.}
    \label{fig:fool-humans}
\end{figure}
}
\fi
%%%%%%%%%%

\end{document}